\documentclass[preprint,12pt]{elsarticle}

\usepackage{amssymb}

\usepackage{amsthm,amsmath}

\usepackage{xcolor}         
\usepackage{bm}
\usepackage{multirow}\usepackage{makecell}
\usepackage{array}

\usepackage{graphicx}

\usepackage{hyperref}
\usepackage{orcidlink}
\usepackage{bm}
\usepackage{cite}
\usepackage{algorithm}
\usepackage{multirow}
\usepackage{color}
\usepackage{xcolor}

\journal{Expert Systems with Applications}

\begin{document}

\begin{frontmatter}

\title{Your Data Is Not Perfect: Towards Cross-Domain Out-of-Distribution Detection in Class-Imbalanced Data}

\author[label1,label2,label3]{Xiang Fang (Corresponding author, Email: xiang003@e.ntu.edu.sg)}

\author[label2,label3]{Arvind Easwaran (Email: arvinde@ntu.edu.sg)}

\author[label3]{Blaise Genest (Email: blaise.genest@cnrsatcreate.sg)}

\author[label4]{Ponnuthurai Nagaratnam Suganthan (Email: p.n.suganthan@qu.edu.qa)}
\affiliation[label1]{organization={Interdisciplinary Graduate Programme-Energy Research Institute @ NTU, Nanyang Technological University},
            country={Singapore}}

\affiliation[label2]{organization={College of Computing and Data Science, Nanyang Technological University},
            country={Singapore}}

\affiliation[label3]{organization={CNRS and CNRS@CREATE, IPAL IRL 2955},
            country={France and Singapore}}
            
\affiliation[label4]{organization={KINDI Computing Research Center, College of Engineering, Qatar University, Doha},
            country={Qatar}}

\begin{abstract}
{Out-of-distribution  detection (OOD detection) aims to detect  test samples drawn from a distribution that is different from the training distribution, in order to prevent models trained on in-distribution (ID) data from providing unavailable outputs. 
 Current OOD detection systems typically refer to a single-domain class-balanced assumption that both the training and testing sets belong to the same domain and each class has the same size. 
 Unfortunately, most real-world datasets  contain 
multiple domains and class-imbalanced distributions,
which severely limits the applicability of existing works. Previous OOD detection systems only focus on the {\em semantic} gap between ID and OOD samples.  
Besides the {\em semantic} gap,
we are faced with two additional gaps: the {\em domain} gap between source and target domains, and the {\em class-imbalance} gap between different classes. In fact, similar objects from different domains should belong to the same class.
In this paper, we introduce a realistic yet challenging setting: 
class-imbalanced cross-domain OOD detection (CCOD), which contains a well-labeled (but usually small) source set for training and conducts OOD detection on an 
 unlabeled (but usually larger) target set for testing. We do not assume that the target domain contains only OOD classes or that it is class-balanced: the distribution among classes of the target dataset need not be the same as the source dataset.
To tackle this challenging setting with an OOD detection system, we propose a novel uncertainty-aware adaptive semantic alignment (UASA) network based on a prototype-based alignment strategy. Specifically, we first build label-driven prototypes in the source domain and utilize these prototypes for target classification to close the domain gap. 
Rather than utilizing fixed thresholds for OOD detection,
we generate adaptive sample-wise thresholds to handle the semantic gap. Finally, we conduct uncertainty-aware clustering to group semantically similar target samples to relieve the class-imbalance gap.
Extensive experiments on three challenging benchmarks (Office-Home, VisDA-C and DomainNet) demonstrate that our proposed UASA outperforms state-of-the-art methods by a large margin.}

\end{abstract}


\begin{keyword}
Out-of-distribution detection \sep Multi-domain alignment \sep Class-imbalanced data \sep Label-driven prototype building \sep Prototype-guided domain alignment \sep  Adaptive threshold generation  \sep Uncertainty-aware target clustering

\end{keyword}

\end{frontmatter}

\begin{figure}[t!]
    \centering
    \includegraphics[width=\textwidth]{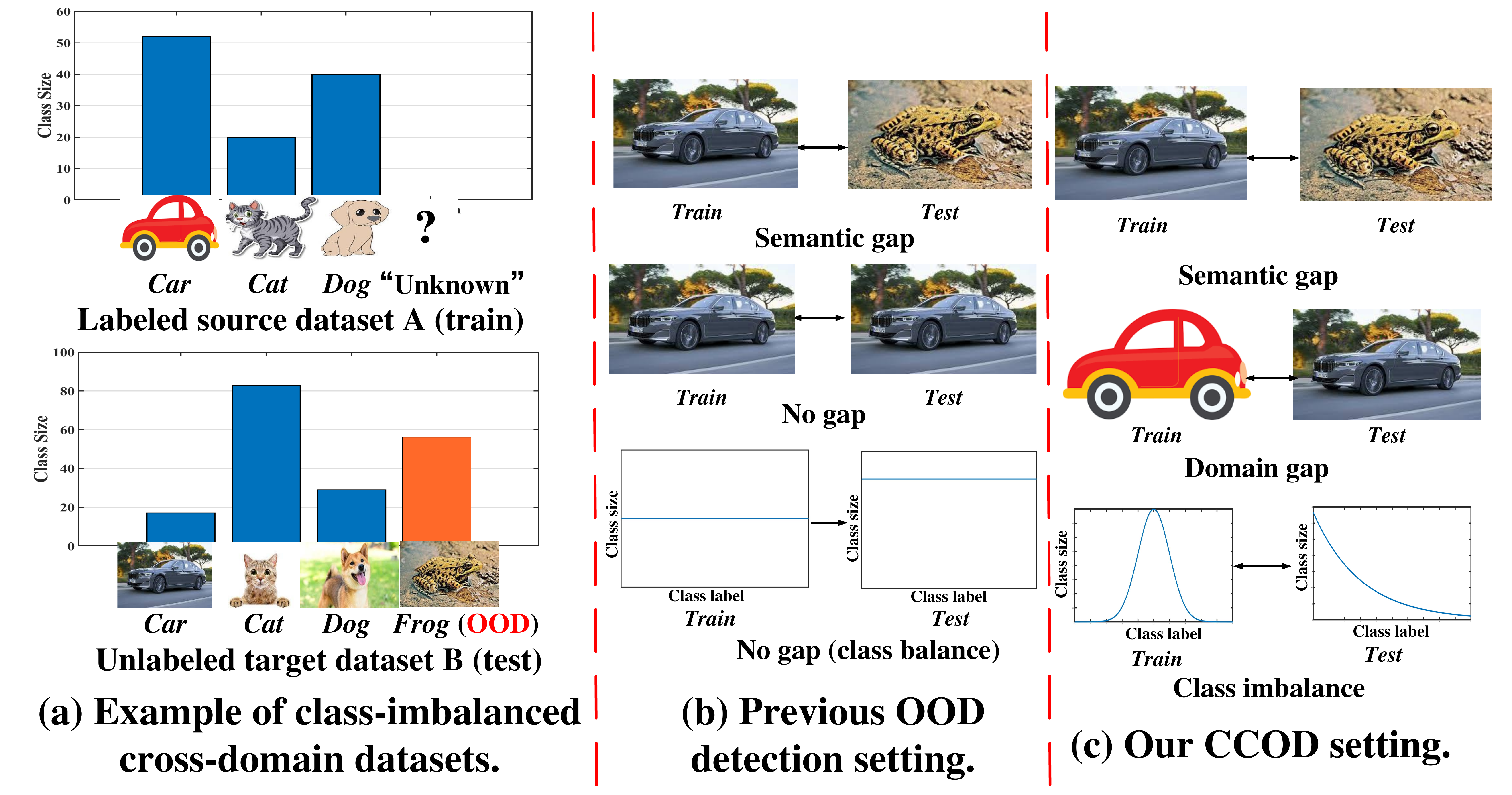}
    \caption{{(a) Example of the class-imbalanced cross-domain out-of-distribution detection (CCOD) setting. (b) and (c) Comparison between previous  models and our proposed model; previous OOD detection methods only address the semantic gap within a single domain while we aim to relieve three gaps: semantic gap (between ID samples and OOD samples), domain gap (between the labeled source domain and unlabeled target domain) and class-imbalance (between different classes). }}
    \label{fig:intro}
\end{figure}
\section{Introduction}
{With the successful development of deep learning, deep neural networks (DNNs) \citep{reimers2020determining,de2021impact,vellido1999neural,yahia2000rough} have been widely applied to many expert and intelligent systems \citep{atmakuru2024deep,maqsood2024mox,wang2025slbdetection} based on a closed-set assumption  that all the test samples are known during training \citep{luo2024dynamic,jiao2024open,li2024auto,neal2018open,cho2022towards,liu2023exploring,wang2025taylor,fang2026towardsicml,kuai2026dynamic,wang2025point,fang2020double,zhang2025monoattack,fang2023hierarchical,liu2024towards,yang2025eood,fang2022multi,fang2026cogniVerse,lei2025exploring,fang2023you,wang2025dypolyseg,fang2025hierarchical,yan2026fit,fang2025adaptive,wang2026topadapter,cai2025imperceptible,fang2026slap,wang2026reasoning,fang2026immuno,wang2026biologically,fang2026disentangling,wang2025reducing,fang2026advancing,fang2026unveiling,wang2026from,liu2023conditional,liu2026attacking,fang2026rethinking,wang2025seeing,fang2026towards,fang2025multi,fang2024fewer,liu2024pandora,fang2024multi,fang2025turing,fang2024not,liu2023hypotheses,fang2024rethinking,liu2024unsupervised,fang2023annotations,xiong2024rethinking,fang2021unbalanced,wang2025prototype,zhang2025manipulating,fang2026align,tang2024reparameterization,fang2025adaptivetai,tang2025simplification,fang2021animc,cai2026towards,fang2020v}. { Unfortunately, real-world datasets contain many outliers that are difficult to distinguish. These outliers are called as out-of-distribution (OOD)  samples, while these non-outliers are treated as in-distribution (ID) samples.
In fact, 
standard DNN-based  systems compulsorily classify both ID and OOD  samples as belonging to one of the known classes \citep{yao2024open,rastegari2016xnor,zhang2022tip,du2020fine}. The wrong classification of outliers will result in irrecoverable losses in some  safety-critical systems, such as autonomous driving \citep{zendel2022unifying,vyas2018out,lu2023uncertainty} and medical diagnosis \citep{ren2019likelihood,zhou2021step}. To solve the above problem,  OOD detection \citep{hendrycksbaseline,sun2022dice,fort2021exploring} is proposed to accurately detect the  outliers  and correctly distinguish the samples from ID classes during testing.}}

{ The largest challenge for an OOD detection system is that no information about OOD samples is available during training, making it difficult to distinguish ID and OOD samples. To  address the challenge, many  OOD detection systems \citep{liang2018enhancing,liu2020energy,sun2021react,lee2018simple,mohseni2020self,vyas2018out} calibrate the distribution of the softmax layer for OOD detection. Other systems \citep{yu2019unsupervised,zaeemzadeh2021out,hsu2020generalized,ming2023exploit} aim to leverage a large number of OOD samples to learn the discrepancy between ID/OOD samples at  training time, then detect the OOD samples during testing. Most OOD detection systems \citep{liang2018enhancing,liu2020energy,sun2021react,lee2018simple,mohseni2020self,vyas2018out} have achieved remarkable performance when the training and testing sets are class-balanced and belong to the same domain. 
In real-world expert systems, the training set and the testing set are often from different domains since there are various difficulties collecting samples from different domains. 
As shown in Figure \ref{fig:intro},
when we train an OOD detection system in the source dataset A, the system can extract knowledge about three ID classes ({\em car}, {\em cat} and {\em dog}) with the same class size in the cartoon domain. When we utilize the trained system on the target dataset B for OOD detection, the model will encounter  samples from an ``unknown'' class (here images of \textit{frog}, not present in the source dataset), which do not belong to the label set of the source domain, and those should be treated as OOD. 
As shown in Figure \ref{fig:intro}(b), the number of samples in any class is nearly the same between the source and the target dataset, \textit{i.e.,  balanced classes}.
Obviously, it is difficult for this assumption to hold true, since it is unrealistic that we collect all the samples from a single domain and make all classes the same size. 
However, domain gap and class-imbalance (Figure \ref{fig:intro}(a)) in real-world OOD detection problems will severely limit their detection ability.}

{ To this end, in this paper we pose a more realistic yet challenging setting: class-imbalanced cross-domain OOD detection (CCOD) to handle the three gaps in Figure \ref{fig:intro}(c): the semantic gap between ID and OOD samples, the domain gap between source and target domains, and the class-imbalance between different classes. The main challenges of our CCOD task are two-fold: 1) a robust OOD detection system should both accurately detect OOD samples and correctly classify ID samples; 2) the designed OOD detection system needs to extract robust and  discriminative features despite the semantic gap, domain gap,  and class-imbalance. To address the aforementioned challenges, we  propose a novel uncertainty-aware adaptive semantic alignment (\textbf{UASA}) network based on four novel and carefully-designed modules: a label-driven prototype building module, a prototype-guided domain alignment module, an adaptive threshold generation module and an uncertainty-aware target clustering module.
In the label-driven prototype building module, we build label-driven prototypes by classifying labeled source samples, where prototypes and labels are bijectively mapped in the source domain.
In the prototype-guided domain alignment module, we leverage these label-driven prototypes to conduct the target classification task.
{ Given a set of ID classes, some OOD samples are  far from all ID samples in the latent space, while others may be semantically close to certain ID classes. Thus, a specific threshold is required for each sample for OOD detection.
For each target sample, we generate an adaptive OOD threshold  to handle the semantic gap in the adaptive threshold generation module. }
Finally, we group the semantically similar target samples into a cluster in the uncertainty-aware target clustering module to reduce the negative impact of class-imbalance.}

{ To this end, our main contributions are summarized as follows:}
\begin{itemize}
  \item {We introduce a more practical and challenging  setting for the OOD detection task called CCOD, where three gaps (semantic gap between ID samples and OOD samples, domain gap between source domain and target domain, and class-imbalance between different classes) are considered. To the best of our knowledge, this is the first attempt to handle all three gaps for OOD detection. }
  \item 
  { For our CCOD task, we propose a novel UASA network with four  modules to handle these challenging gaps. Specifically, to close the domain gap, we build label-driven prototypes in the source domain and leverage these prototypes for classification in the target domain. To handle  the semantic gap for each target sample, we  generate an adaptive threshold for OOD detection. As for the class-imbalance, we conduct uncertainty-aware clustering to align target samples that share similar semantics.}
  \item { Extensive experimental results on three challenging class-imbalanced benchmarks (Office-Home, VisDA-C and DomainNet) demonstrate that  UASA  outperforms existing state-of-the-art approaches by a large margin. In representative cases, UASA beats all compared methods by 9.06\% on the DomainNet dataset. }
\end{itemize}

{ The rest of this paper is organized as follows: Section \ref{re} presents related work. Section \ref{me} describes  our proposed UASA network for  our CCOD task. Section \ref{ex} illustrates our performance on three challenging benchmarks. Section \ref{con} concludes this paper.}

\section{Related Work}
\label{re}
{\noindent \textbf{Out-of-distribution detection}.
As a challenging expert system task, OOD detection  aims to detect test samples from distributions that do not overlap with the training distribution. Previous OOD detection systems \citep{liang2018enhancing,liu2020energy,sun2021react,lee2018simple,mohseni2020self,vyas2018out,yu2019unsupervised,zaeemzadeh2021out,hsu2020generalized,ming2023exploit}  can be divided into four types: classification-based systems \citep{hendrycksbaseline,liang2018enhancing,lee2018hierarchical,lee2018training}, density-based systems \citep{kirichenko2020normalizing,serrainput}, distance-based systems \citep{techapanurak2020hyperparameter,lee2018simple} and reconstruction-based systems \citep{zhou2022rethinking,yang2022out}. Although previous systems have achieved decent success, most of them assume that all the datasets are from the same domain and all the classes are balanced. 
In fact, we always collect training and testing sets from various domains, and each class has a unique number of samples.
Unlike the aforementioned OOD detection systems that only address the semantic gap between ID and OOD samples, our proposed system targets to handle three challenging gaps: semantic gap between ID and OOD samples, domain gap between source and target domains and class-imbalance between different classes.}

{\noindent \textbf{Unsupervised domain adaptation (UDA).}
As a significant technology in expert systems, UDA \citep{ganin2015unsupervised,kang2019contrastive,long2016unsupervised,ru2023imbalanced} aims to transfer predictive systems trained on fully-labeled data from a source domain to an unlabeled target domain. The primary objective of existing UDA systems, which are predominantly classification-based, is to mitigate the domain gap between the source and target domains \citep{damodaran2018deepjdot,ganin2015unsupervised,long2015learning,long2017deep,tzeng2014deep}. By effectively aligning the statistical distributions of the source and target domains, these systems strive to enhance the generalization capability of the model in the target domain.
Furthermore, the realm of UDA has witnessed significant advancements in the domain of visual tasks, such as video action recognition \citep{chen2019temporal,choi2020shuffle,munro2020multi} and video segmentation \citep{chen2020action,ullah2024video,fang2024source}. 
UDA-based systems \citep{zhang2024cross,liu2024learning,cui2024unified} have been successfully extended to these visual tasks, enabling knowledge transfer from the labeled source domain to the unlabeled target domain, thus circumventing the need for costly manual annotation in the target domain. Previous UDA-based systems only refer to the closed setting where all the test samples are ID \citep{zhang2020unsupervised,ainam2021unsupervised,yang2022unsupervised}. Therefore, these UDA-based systems under the closed setting are not applicable to the challenging CCOD task. Unlike them, our proposed UASA can handle the CCOD task with four carefully-designed modules.}

{\noindent \textbf{Class-imbalanced domain adaptation (CDA)}. The CDA task \citep{tan2020class,yang2022multi,prabhu2021sentry} is a branch of domain adaptation, which aims to carry out domain alignment on multi-domain expert systems with biased class distribution. The main challenge for the CDA task is the class-imbalance gap  \citep{tachet2020domain,proto_uda}.  
To relieve the class-imbalance gap, many works conduct domain adaptation on class-imbalanced datasets by exploiting the pseudo-labeled target samples \citep{tan2020class} or utilizing a sample selection strategy \citep{jiang2020implicit}.
Although the above systems can close the class-imbalance gap, they focus on the ID classification task and cannot deal with the OOD detection task, which limits their applications in real-world open-set datasets. Unlike them, we can conduct ID classification in the prototype-guided domain alignment module and detect OOD samples with the adaptive threshold generation module.}

\begin{figure*}[t]
  \centering
   \includegraphics[width=\textwidth]{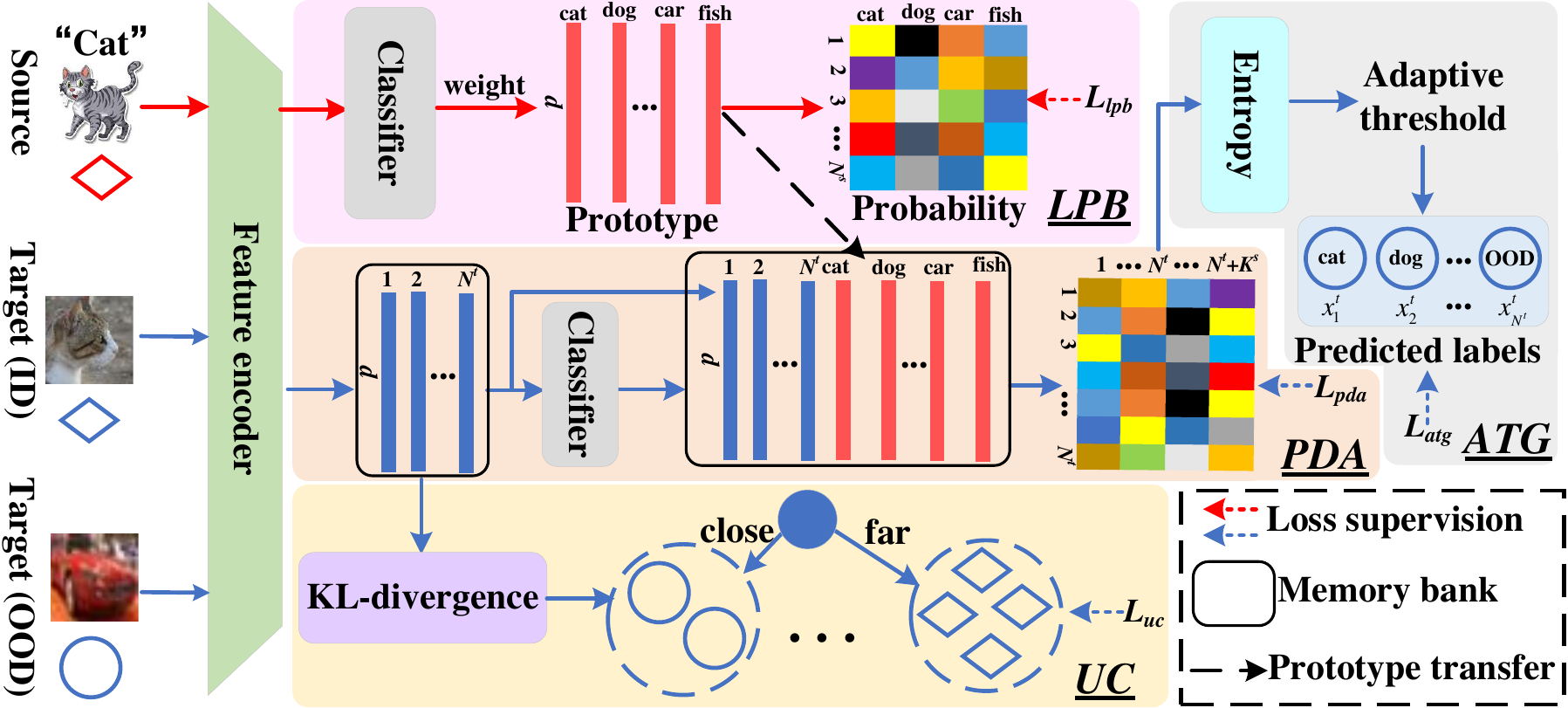}
   \caption{
   {Overview of our proposed UASA system  for the proposed CCOD task.} \underline{Underlined module names} refer to the eponymous sections in the text. First, we feed all source and target images into a ResNet-50 network \citep{resnet} to extract their features. (i) In the label-driven prototype building (\underline{\textit{LPB}}) module, we build  label-driven source prototypes by classifying source images. (ii) In the prototype-guided domain alignment (\underline{\textit{PDA}}) module, we leverage these prototypes in a memory bank for target classification. (iii) In the adaptive threshold generation (\underline{\textit{ATG}}) module, we automatically generate a threshold for each target sample to distinguish if the sample is OOD or ID. If it is ID, we choose the label with the highest probability as its label; otherwise, we mark it as OOD. (iv) In the uncertainty-aware target clustering (\underline{\textit{UC}}) module, we cluster target samples into different clusters. By aligning semantically similar samples in each cluster, we alleviate the class-imbalance gap.}
   \label{method}
\end{figure*}

\section{The CCOD Task and Our Proposed UASA}
\label{me}
{\noindent \textbf{Task definition for CCOD.}
Given a source set $D^s=\{(x_i^s,y_i^s)\}_{i=1}^{N^s}$ with $N^s$ samples 
$\{x_i^s\}_{i=1}^{N^s}$ together with associated class-labels $\{y_i^s\}_{i=1}^{N^s}$,
and a target domain $D^t=\{x_j^t\}_{j=1}^{N^t}$ with $N^t$ samples $s_i^t$, our posed CCOD task aims to train an OOD detection system on the combined domain $D^s \cup D^t$ to correctly classify the target samples into one of the  classes  shared with source domain and group OOD samples into $K^t$ OOD classes, where $K^t \geq 1$.
If the source label set is $Y^s=\{1,2,...,K^s\}$, the predicted target label is $Y=Y^t=\{1,...,K^s,K^s+1,...,K^s+K^t\}$, where the last $K^t$ classes are unique to the target domain, \textit{i.e.}, OOD classes.
{In real-world multi-domain datasets, different classes contain various numbers of samples, \textit{i.e.}, these classes are imbalanced.
For the source  and target domains, previous domain adaptation works \citep{turrisi2022multi,nguyen2021most} usually refer to the {\em class-balanced} assumption, that is: 
$p^s(x|y=c) = p^t(x|y=c), \forall c \in Y^s$, where $p^s$ and $p^t$ denote the probability density functions of the source and target distributions, respectively.
In our CCOD task, such a strict assumption is not imposed and each class can contain any number of samples, \textit{i.e.}, $p^s(x|y=c) \neq p^t(x|y=c), \exists c \in Y^s$.}}

{\noindent \textbf{UASA network.}
To address the CCOD task, we propose a novel uncertainty-aware adaptive semantic alignment (UASA) system  in Figure \ref{method}.}
{Specifically, to close the domain gap between source and target domains, we  design a  prototypical network to align source and target samples with the same semantics. To handle the semantic gap between ID and OOD samples, we propose an  adaptive threshold generation module to  distinguish ID and OOD samples in a gradual and fine-grained manner. To relieve the class-imbalance gap between different classes in the target domain, we design an uncertainty-aware target clustering strategy to form tight  clusters with semantically similar samples.}

\subsection{Label-Driven Prototype Building for Source Classification} \label{source_classification}
To fully understand these images, the designed system is required to close the domain gap between  source  and  target domains.
{Therefore, most UDA-based systems \citep{peng2020domain2vec,melas2021pixmatch,kang2019contrastive} conduct domain alignment by mapping  source  and target features into a shared latent space. Although the feature mapping strategy can close the domain gap to a certain extent, it has the following disadvantages}: 1) It is sensitive to OOD samples. Directly utilizing source samples will limit the generalization ability of the designed model, which makes it difficult for the model to distinguish between the domain gap and the semantic gap. { 2) In a  mini-batch, missing classes from the source domain  will lead to inappropriate domain adaptation, 
and missing classes cannot contribute to domain adaptation, limiting their performance. 3) The mapping strategy is sensitive to the class size, since UDA-based  systems always pay more attention to the classes with more samples and ignore the small-scale classes, which will limit the generalization ability of the designed system.}

{ To transfer the classification knowledge from the source domain to the target domain, we aim to construct a label-driven prototype based on the labeled source images.
We first utilize a ResNet-50 network \citep{resnet} to extract the image features in the source domain. 
Then, a fully-connected layer with $\mathcal{L}_2$-normalization serves as the classifier, where  the learnable classifier weights are denoted as $M = \{m_c\}_{c=1}^{K^s} \in \mathbb{R} ^ {d \times {K^s}}$, where $m_c$ is the weight of the $c$-th class and $d$ denotes the feature dimension.
Finally, we learn the consensus representation of each class based on the weight matrix $M$.
For any  sample  $x_i^s \in \mathbb{R} ^ d$ in the source domain and its corresponding label $y_i^s\in \mathbb{R}$, we can train the source classifier
with the following cross-entropy loss:}
\begin{equation}
  \label{l_ce}
  \mathcal{L}_{lpb} = - \frac{1}{N^sK^s} \sum_{(x_i^s,y_i^s)\in D^s} \sum_{c=1}^{K^s}\log {{p}}^{s}(y_i^s=c|x_i^s,m_c) \cdot \bm{1}_{\{y_i^s=c\}},
\end{equation}
where ${{p}}^{s}(y_i^s=c|x_i^s,m_c)$ is the predicted probability that $x_i^s$ is in class $c$, which is computed as: 
\begin{equation}
  \label{temp}
  {{p}}^{s}(y_i^s=c|x_i^s,m_c)= \frac{e^{(1/\sigma\cdot{m}_c^{\top}x_i^s)}}{\sum_{j=1}^{K^t}e^{(1/\sigma\cdot{m}_j^{\top}x_i^s)}},
\end{equation}
where the temperature parameter $\sigma$ balances the concentration degree \citep{tao}. Note that $m_j$ is class-specific, \textit{i.e.}, each column of $M$ corresponds to a unique class.
Therefore, we treat $m_j$ as the source prototype of the $j$-th class.  The prototypes we utilize have four significant advantages over those in previous works \citep{gao2024learning,belal2024multi,cui2024effective}: 1) They are robust to OOD samples, which can effectively assist our proposed system with domain alignment. 2) Since the weights are class-specific, these prototypes can serve as the representation of each class during training. 3) Some real-world expert systems deal with highly private data. Previous systems \citep{fang2024prototype,sun2025prototype,xie2024adapting} needed to access all the data to build their prototypes, however, our prototypes allow our system to conduct domain adaptation without accessing the source dataset, which preserves data privacy in the source domain. 4) We build the prototypes without introducing  additional parameters, which significantly reduces the computational cost to obtain the prototypes.

\subsection{Prototype-Guided Domain Alignment for Target Classification}
{ To transfer knowledge from the labeled source domain to the unlabeled target domain, previous UDA-based systems \citep{nguyen2022cycle,sicilia2022pac,ren2024towards} mitigate the domain gap between the source and target domains in an adversarial way. 
However, the real-world setting is more challenging due to the semantic gap (between ID and OOD samples) and imbalanced  classes.
Therefore,  forceful domain alignment between the source domain and the target domain might lead to catastrophic misalignment. To this end, we design a prototype-guided domain alignment module for target classification.
To correctly align semantically similar samples, we feed all $\mathcal{L}_2$-normalized target features into a memory bank ${{Z}}=[{z}_1, {z}_2,...,{z}_{N^t}] \in \mathbb{R} ^ {d \times N^t}$. Then, we concatenate the memory bank $Z$ and the source classification weight $M$ to obtain the cross-domain representation $F=[Z,M]=[{z}_1,...,{z}_{N^t}, {m}_1,...,{m}_{K^s}]$, where $[\cdot,\cdot]$ denotes the concatenation operation.
During each iteration, we update the target features in the memory bank ${{Z}}$. Therefore, we use the memory bank to conduct domain alignment with the following   loss:}
\begin{equation}
  \label{l_nc}
  \mathcal{L}_{pda}=-\frac{1}{|\mathcal{B}^t|(N^t+{K^s})}\sum_{i \in \mathcal{B}^t}\sum_{j=1,j \ne i}^{N^t+{K^s}}p^t_{ij}{\log}(p^t_{ij}),
\end{equation}
{where $p_{ij}^t={e^{(1/\sigma \cdot {{f}}_j^{\top}x_i^t)}}/{\sum_{l=1,l \ne i}^{N^t+{K^s}}e^{(1/\sigma \cdot{{f}}_l^{\top}x_i^t)}}$ denotes the probability that the target sample $i$ belongs to label $y_j$}; $\mathcal{B}^t$ is the corresponding mini-batch of each target domain mini-batch;  and ${{f}}_j$ and ${{f}}_l$ are the $j$-th and $l$-th columns of $F$, respectively.
We can minimize the entropy of the similarity between the target samples and source prototypes under Equation \eqref{l_nc}, which aligns each target sample to a source prototype or its paired neighbor in a mini-batch. { Therefore, these target ID samples will be classified into the correct classes. }

\subsection{Adaptive Threshold Generation for Target OOD Detection}
\label{atg}
{ The main challenge of our CCOD task is correctly detecting OOD samples in the target domain. 
In real-world systems,  
the relationship between ID and OOD samples varies significantly. For example, some OOD samples are  far from all ID samples in the latent space and are easy to recognize as OOD samples. For convenience,  we refer to these OOD samples as  ``easy OOD samples''. On the contrary, some OOD samples may be semantically close to certain ID classes, and we call them ``hard OOD samples'' as they are hard to distinguish as OOD samples.
During inference, OOD detection systems tend to overconfidently predict the results for those hard OOD samples as similar to ID classes, which corresponds to a small threshold for the similarity to ID samples. Also, some classes may contain more easy OOD samples, where we need a larger threshold to avoid OOD samples being misclassified as ID classes. Considering that easy OOD samples have higher entropy, a higher threshold can help us detect these OOD samples and obtain better performance.
We observe that in most OOD detection systems, ID samples always have lower entropy than OOD samples \citep{li2020background,sun2022exploiting,vernekar2019out}, which is an essential criterion to distinguish ID and OOD samples.
Previous OOD detection systems predefine a fixed entropy threshold $o'$ for all the samples to assess whether any target sample $x_i^t$ is OOD or not. If the entropy of $x_i^t$ is greater than $o'$, $x_i^t$ is OOD, and vice versa. 
In fact, on real-word datasets we need different thresholds to handle various scenarios. The fixed  threshold  makes previous OOD detection systems  sensitive to parameter tuning and weakens their robustness in complex  scenarios. Therefore, we design a novel adaptive threshold generation policy to learn an adaptive threshold for each target sample.}

{ To adaptively generate sample-wise thresholds in the target domain, we generate a pseudo-label for each target sample in a self-supervised manner. { For convenience, $H_i$ denotes the set of target samples that are labeled as class $i$.} Initially, we do not distinguish OOD samples, \textit{i.e.}, $H_1  \cup ... \cup H_{K^s}=\{1,...,N^t\}$. Since the easy OOD samples often have larger entropy than the hard OOD samples, we calculate the  mean entropy of each class to recognize which class consists of more easy OOD samples.
We denote the softmax output of the classifier as ${{p^t}}$, where ${{p}_i^t}$ denotes the $i$-th target sample's probability distribution. Thus, its corresponding class-wise entropy $T_i$ is given by: }
\begin{equation}
  T_i=\frac{1}{|H_i|}\sum_{j=1}^{|H_i|}Q({{p}}^t_{H_{ij}}),
\end{equation}
where $i \in Y^t$; and $Q(\cdot)$ is the sample-wise entropy function.
We set the threshold $o_i$ of the target sample $x_i^t$ using its entropy $T_i$ by:
\begin{equation}
  \label{q}
  o_i=\frac{\alpha({T}_i- \min({{T}})- \max({{T}}))}{\max({{T}}) - \min({{T}})} \log K^s,
\end{equation}
{where $\alpha$ is an adjustable hyperparameter.
Based on the sample-wise threshold $o_i$, we classify ID samples and detect OOD samples. Thus, for the target sample $x_i^t$, its predicted label is:}
\begin{equation}
  \bar{y}_{i}^{t} =
  \begin{cases}
  \mathop{\arg\max}\limits_{j} {{p}^t_{ij}}, & Q({{p}}^t_i) \leq o_{\mathop{\arg\max}\limits_{j} {{p}^t_{ij}}},\\
    OOD, & \text{otherwise,}
  \end{cases}
\end{equation}
where the adaptive threshold $o_{\mathop{\arg\max}\limits_{j} {{p}^t_{ij}}}$ will be different for individual target samples.
If  entropy $Q({{p}}^t_i)$ is not higher than the adaptive threshold, $x_i^t$ is ID and we predict its label as  $\mathop{\arg\max}\limits_{j} {{p}^t_{ij}}$. Otherwise, $x_i^t$ is OOD.
 To obtain a clearer ID/OOD decision boundary, we utilize the following entropy separation loss:
\begin{align}
\label{l_es}
  \mathcal{L}_{atg}&=-\frac{1}{|\mathcal{B}^t|}\sum_{i \in \mathcal{B}^t}\mathcal{L}_{atg}({{p}}^t_i),\nonumber\\
  \mathcal{L}_{atg}({{p}}^t_i)&=
  \begin{cases}
  0, & ||o'_i-Q({{p}}^t_i)||_2^2 < \Delta, \\
    ||o'_i-Q({{p}}^t_i)||_2^2, & \text{otherwise,}
  \end{cases}
\end{align}
where $||\cdot||_2$ denotes the $\mathcal{L}_2$ norm,
$o'_i=o_{\mathop{\arg\max}\limits_{j} {{p}^t_{ij}}}$, and $\Delta$ is the confidence interval.
The confidence interval $\Delta$ allows us enable the entropy separation loss to keep target samples from the ID/OOD decision boundary.

\subsection{Uncertainty-Aware Target Clustering for Class-Imbalance}
{To close the class-imbalance gap, we design a novel uncertainty-aware clustering module for the target domain. }
During clustering, we first align the target samples with their corresponding cluster centers by conducting clustering (\textit{e.g.}, K-means clustering) and then tightening the clusters. In a target domain mini-batch $\mathcal{B}^t$, we denote any two samples as $x_{i}^{t}$ and $x_{j}^{t}$, and their corresponding probability distributions as  ${{p}_{i}^t}$ and ${{p}_{j}^t}$, respectively.
If $x_{i}^{t}$ and $x_{j}^{t}$ share the same predicted label, we should align them. For supervision, we introduce the Kullback Leibler divergence loss (KL loss):
\begin{equation}
  \label{l_kl}
  \mathcal{L}_{kl}({{p}_{i}^t}, {{p}_{j}^t}) = \frac{1}{2} [\mathcal{F}_{kl}({{p}_{i}^t} | {{p}_{j}^t}) + \mathcal{F}_{kl}({{p}_{j}^t} | {{p}_{i}^t})], 
\end{equation}
{where $\mathcal{F}_{kl}(\cdot|\cdot)$ denotes the KL divergence function between two samples. Since real-world OOD samples are often semantically different, directly minimizing the distance between them will confuse the OOD detector, resulting in inaccurate OOD detection results. To only align semantically samples, we feed all $\mathcal{L}_2$-normalized target features into a memory bank $Z=[{z}_1, {z}_2,...,{z}_{N^t}] \in \mathbb{R} ^ {d \times N^t}$. During each epoch, we update the target features in the memory bank ${Z}$.  Then, we conduct uncertainty-aware target clustering on the memory bank $Z$ to obtain $A$ clusters, where ${C} \in \mathbb{R} ^ {d \times A}$ denotes these clusters and $A>K^s$.} We can obtain the corresponding cluster index for each target sample $x_j^{t}$ as follows:
\begin{equation}
  a_j = \mathop{\arg\max}\limits_{i} {\cos(c_{i},x_j^{t})},
\end{equation}
{where $\cos(\cdot,\cdot)$ denotes the cosine similarity function, and $c_{i}$ is the $i$-th column of ${C}$, which  can be treated as the  center of the $i$-th cluster. Based on the clustering strategy, semantically different samples will be allocated to different clusters. For convenience, we define the set of pseudo-labels  as $\{1,\cdots,K^s,\cdots,A\}$. }

Incorrect pseudo-labels may lead to unsatisfactory OOD detection performance if we directly utilize these labels in the KL loss. To relieve the negative effect of incorrect pseudo-labels, we define any two samples $(x_{i}^{t}$ and $x_{j}^{t})$ in the same mini-batch as a pair $\{x_{i}^{t}, x_{j}^{t}\}$.
Therefore, we should assign different weights to different pairs.
In particular, we will assign higher weights to the pairs with higher confidence scores, and vice versa.
If a target sample is labeled as an ID sample in Section \ref{atg}, we use the largest logit of its probability distribution across ID classes as its confidence score. For an OOD sample, we define its confidence score as the entropy of its probability distribution.
In the mini-batch, we define the weight $m_{ij}$ of each pair $\{x_{i}^{t}, x_{j}^{t}\}$ as:
\begin{equation}
  m_{ij}=
  \begin{cases}
  0, & \bar{y}_{i}^{t} \neq \bar{y}_{j}^{t},\\ 
    \frac{s_i+s_j}{2}, & \text{otherwise,} 
  \end{cases}
\end{equation}
where $s_i$ and $s_j$ are confidence score of target samples $i$ and $j$, respectively.
{ Therefore, we introduce the following loss to conduct uncertainty-aware target clustering:}
\begin{equation}
  \label{l_cl}
  \mathcal{L}_{uc}=\frac{1}{|\mathcal{B}^t|} \sum_{i \in \mathcal{B}^t,j \in \mathcal{B}^t, i \ne j} m_{ij} \cdot \mathcal{L}_{kl}({{p}_{i}^t}, {{p}_{j}^t}).
\end{equation}
{ In Equation \eqref{l_cl} we make  semantically similar samples (\textit{i.e.}, samples that share the same pseudo-label) tighter, which will  improve the robustness of our system under the class-imbalance gap.}

Overall, our total loss is formulated as follows:
\begin{equation}
  \label{all}
  \mathcal{L} = \mathcal{L}_{lpb} + \lambda_1 \mathcal{L}_{pda} + \lambda_2\mathcal{L}_{atg} + \lambda_3 \mathcal{L}_{uc},
\end{equation}
where $\lambda_1$, $\lambda_2$ and $\lambda_3$ are hyperparameters that balance the importance of the different losses.

\section{{Experiments and Analysis}}
\label{ex}

\noindent \textbf{Datasets.} 
{ To evaluate the performance of our proposed system, we need multi-domain datasets.}
Following \citep{saito2021ovanet}, we utilize three popular yet challenging datasets: DomainNet \citep{domainnet}, Office-Home \citep{oh} and VisDA-C \citep{visda,isfda}. 

{1) DomainNet \citep{domainnet,coal} contains 600k images from 345 classes on 4 widely used domains: real (\textbf{R}), clipart (\textbf{C}), painting (\textbf{P}) and sketch (\textbf{S}). All 345 classes are present in each domain. We choose a domain as the labeled source domain and the remaining three domains as unlabeled target domains. By treating one domain as the source domain and another domain as the target domain, we can construct a transfer task from the source domain to the target domain.
Considering all the domains, we track the following transfer tasks: {R$\rightarrow$C}, {R$\rightarrow$S}, {R$\rightarrow$P}, {C$\rightarrow$R}, {C$\rightarrow$S}, {C$\rightarrow$P}, {S$\rightarrow$R}, {S$\rightarrow$C}, {S$\rightarrow$P}, {P$\rightarrow$C}, {P$\rightarrow$S}, {P$\rightarrow$R}. In each source domain,  we regard the first 45 classes in alphabetical order as the source domain input, and drop the remaining (300 classes). In each target domain, we utilize all 345 classes  as the target domain input. For example, in the  R$\rightarrow$C task, we use the first 45 classes in the R domain as the source domain input, and all 345 classes (the first 45 classes are ID and the last 300 classes are OOD) in the C domain as the target domain input.}

2) Office-Home \citep{venkateswara2017deep} contains 15,500 images from 65 classes on 4 domains: art (AR), clipart (CL), product (PR) and real-world (RE). Due to the page limitation, we only utilize 3 domains: CL, PR and RE. Similarly, we select one domain as the labeled source domain and the remaining two domains as unlabeled target domains.
We consider 6 transfer tasks: {RE$\rightarrow$PR}, {RE$\rightarrow$CL}, {PR$\rightarrow$RE}, {PR$\rightarrow$CL}, {CL$\rightarrow$RE} and {CL$\rightarrow$PR}.   We choose the first 50 classes in alphabetical order as each source domain's input. In the target domain, all the classes (50 ID classes and 15 OOD classes) are used as input.

{3) Following \citep{isfda}, we utilize the class-imbalanced version of VisDA-C \citep{visda,isfda} for our CCOD task. VisDA-C contains 280,157 images from 12 classes on 3 domains. Similarly, we select a labeled source domain and treat the remaining two domains as unlabeled target domains. We denote the class size of the $i$-th class as $O_i$. For convenience, we define $N_{\max}=\max\{O_1,O_2,...,O_{12}\}$ and $N_{\min}=\min\{O_1,O_2,...,O_{12}\}$ 
 An  imbalance factor $\mu$ is used to indicate the degree of class-imbalance, which is defined by $\mu={N_{\max}}/{N_{\min}}$. }

\noindent \textbf{Implementation details.}
For  a fair comparison, we follow \citep{dance} and use a ResNet-50 network \citep{resnet} as the feature encoder. We replace the last layer of the ResNet-50 network with the new weight matrix to obtain the weight of each class during classification. 
 For all the datasets, we set $\sigma=0.05$ in Equation \eqref{temp},   $\Delta=0.5$ in Equation \eqref{l_es}, $\alpha=0.15$ in Equation \eqref{q}, and $A = 2.5K^s$. In Equation \eqref{all}, we set
$\lambda_1=0.05$, $\lambda_2=0.1$, $\lambda_3=0.1$. We set the mini-batch size $|\mathcal{B}^t|$ as 128 on  VisDA-C, and 32 on Office-Home and DomainNet. Similarly, the learning rate is set at $1\times10^{-2}$ on  Office-Home and DomainNet, and $2\times 10^{-2}$ on  VisDA-C . The stochastic gradient descent optimizer with a momentum of 0.9 is utilized for model optimization. We implement our proposed method in PyTorch \citep{torch}. Our codes are available in \href{https://github.com/AngeloFang/UASA}{Github}.

\begin{table*}[t!]
  \begin{center}
    \caption{ Performance comparison with state-of-the-art methods for  CCOD  on the DomainNet dataset.}
  \scalebox{1}{   \setlength{\tabcolsep}{0.25mm}{
  \begin{tabular}{l|cccccccccccccccccc}
  \hline
  {Models} & {R$\rightarrow$C}       & {R$\rightarrow$S}       & {R$\rightarrow$P}       & {C$\rightarrow$R}       & {C$\rightarrow$S}       & {C$\rightarrow$P} & {S$\rightarrow$R}       & {S$\rightarrow$C}       & {S$\rightarrow$P}       & {P$\rightarrow$C}       & {P$\rightarrow$S}       & {P$\rightarrow$R}    & {Mean }       \\\hline
  MSP& 28.75&30.14&28.76&10.03&24.11&21.76& 12.93&19.54& 20.17& 14.26& 10.76& 11.29& {19.38}\\
  ODIN& 32.96& 33.78& 35.82&16.92&23.70&20.53&18.23&16.75&23.14&13.92&16.75&10.33&{21.90}\\
   UAN  &45.21&50.85&52.30&20.53&32.62&36.81&26.15&19.03&29.14&23.28&15.69&9.43 & {30.09}   \\
Energy&46.72&44.59&51.93&38.46&35.72&34.16&24.83&22.69&31.72&28.94&24.83&49.52 & {36.18}\\
     CMU    &49.12&48.17&52.58&51.36&42.89&40.43&52.13&47.80&48.35&43.86&46.73&48.57&  { 47.67}        \\
     CIDER&50.96&50.37&54.82&48.99&47.32&51.94&55.36&40.83&51.42&44.17&48.32&51.96& {49.71}\\
  ROS &54.17&52.03&56.49&71.24&56.50&59.31&70.52&52.28&60.74&45.59&50.97&57.86 & {57.31}       \\
  STA & 57.53&  57.48& 61.02& 62.19&  59.96& 57.25& 69.04& 59.02& 62.83& 52.33& 52.01& 51.65 & {58.53 }  \\
    DANCE & 60.28 &59.93&65.86&70.82&58.53&60.32&73.77&  62.40&68.19&56.31&59.17&64.28  & {63.32}     \\
  OSBP & 61.35&61.04&67.19&72.63&53.75&54.33&73.05&54.44&67.23&53.14&59.42&66.20   & {61.98 }    \\
  Ovanet &64.01& 63.43&67.65&71.13& 61.42&65.43&70.36&66.75&72.38&62.19&63.94&68.25 & {66.41}\\\hline
  \textbf{UASA} &\textbf{68.53}&\textbf{65.12}&\textbf{69.90}&\textbf{77.82}&\textbf{62.88}&\textbf{65.52}&\textbf{83.46}&\textbf{68.51}&\textbf{73.05}&\textbf{63.23}&\textbf{65.18}&\textbf{77.31}   & {\textbf{70.04} }                       \\ \hline
  \end{tabular}}
  }
  \label{result_dn}
  \end{center}
\end{table*}
\begin{table}[t!]
  \begin{center}
    \caption{Performance comparison with state-of-the-art methods for  CCOD  on the Office-Home dataset.}
    \scalebox{1}{   \setlength{\tabcolsep}{1.8mm}{
    \begin{tabular}{l|ccccccccccccc}
      \hline
{Models}            & {RE$\rightarrow$PR}                                              & {RE$\rightarrow$CL}                                              &{PR$\rightarrow$RE}                                              & {PR$\rightarrow$CL}                                              & {CL$\rightarrow$RE}                                                & {CL$\rightarrow$PR}    & {Mean}     \\\hline
      UAN    &  14.82&6.13&9.52&3.82&  6.17&8.32 & {8.13 }       \\
      MSP & 18.62& 12.35& 11.17&6.28&10.37&12.84 & {11.94}\\
      ODIN& 22.31& 18.30& 10.72& 8.13& 11.55&14.72 & {14.29} \\
      ROS      & 50.26&44.12&54.62&42.87&43.50&35.72 & {45.18 }       \\
       CIDER&50.49&42.13& 50.26& 40.38& 39.75&34.33 & {42.89} \\
      Energy& 51.37&40.19&52.75& 38.26&44.23&37.52 & {44.05} \\
      CMU     & 53.04&39.82&50.76&35.10&41.35&36.29   & {42.73 }      \\
      OSBP    &  60.34&49.12& 63.27&46.20&56.37& 57.43 & {55.46}       \\
      STA     &  61.23&48.01&64.72&48.96& 54.62& 55.34  &{55.48 }  \\
      DANCE  &  61.27& 50.93&63.18&53.96&60.15&59.03 &  {58.09}       \\
      Ovanet   &  63.21&49.02&67.72&46.23&62.07&57.48  & {57.62}       \\
\hline
      \textbf{UASA}     & \textbf{67.48} &\textbf{52.53}&\textbf{67.85}&\textbf{55.29}&\textbf{62.59}&\textbf{61.08} & {\textbf{61.14}} \\ \hline
      \end{tabular}}
    }
    \label{result_oh}
  \end{center}
\end{table}

\begin{table}[t!]
  \begin{center}
  \caption{Performance comparison with state-of-the-art methods for  CCOD on  the VisDA-C dataset.}
  \scalebox{1}{   \setlength{\tabcolsep}{0.4mm}{
  \begin{tabular}{l|ccccccccccccccccccccc}
  \hline
\multirow{2}{*}{Models} & {$\mu=100,$}                                              & {$\mu=50,$}                                               & {$\mu=10,$}    & {$\mu=5,$}                                            & {$\mu=100,$}                                             & {$\mu=50,$}  & {$\mu=10,$} & {$\mu=5,$}  &\multirow{2}{*}{{Mean}}  \\
~& $K^s=9$& $K^s=9$& $K^s=9$& $K^s=9$& $K^s=10$& $K^s=10$& $K^s=10$& $K^s=10$&\\\hline
ODIN&18.52&27.34& 28.17& 28.40& 29.43& 37.92& 34.95& 35.72 & {30.06}\\
  CMU& 23.58& 25.31& 29.04& 29.18& 29.82& 39.37 & 33.86& 35.29 & {30.68} \\
  MSP&25.32& 22.18&20.16& 20.87& 27.92&36.83& 30.24& 33.42 & {27.12}\\
  Energy&32.19&29.43&23.75 &23.52 &20.34&26.17& 23.59& 24.96 & {25.50}\\
   OSBP& 42.85& 43.62& 43.95& 44.03& 45.38& 48.12& 46.73& 45.72 &{45.05} \\ 
  STA& 43.08&43.62&45.37& 45.78&42.14&40.13& 41.09& 41.27 & {42.81} \\
  CIDER&44.18&42.56&43.82& 43.26&40.29&41.13& 42.40& 41.35 & {42.37}\\
  ROS & 45.12& 44.93&49.55& 50.03& 45.27& 40.86& 44.38& 44.79 & {45.62} \\
  Ovanet & 46.40& 48.73& 51.08& 50.84& 47.13& 49.41&47.28& 48.33 & {48.65} \\
  DANCE & 50.13& 44.09& 48.70& 48.62& 41.68& 39.67 & 39.55& 40.18 & {44.08}\\ 
    UAN & 50.21& 51.14& 48.32& 49.36& 43.07& 44.18& 43.86& 43.72 &{46.73}  \\\hline
  \textbf{UASA} & \textbf{54.11}& \textbf{54.96}& \textbf{54.07}& \textbf{54.29} &\textbf{48.03}& \textbf{52.16} & \textbf{50.13}& \textbf{51.34} & {\textbf{52.39}} \\ \hline
  \end{tabular}
  }}
  \label{result_visda}
  \end{center}
\end{table}

\begin{table*}[t!]
  \begin{center}
  \caption{ Performance comparison  for the CCOD task  on the Office-Home dataset based on Saito et al.'s settings\citep{ova,dance}.}
  \scalebox{1}{   \setlength{\tabcolsep}{1.8mm}{
    \begin{tabular}{l|ccccccccccccc}
      \hline
{Models}            & {RE$\rightarrow$PR}                                              & {RE$\rightarrow$CL}                                              &{PR$\rightarrow$RE}                                              & {PR$\rightarrow$CL}                                              & {CL$\rightarrow$RE}                                                & {CL$\rightarrow$PR}   &{Mean}      \\\hline
STA& 53.45 &45.14 &47.12& 41.27& 43.68&43.72 &{45.73}\\
OSBP& 46.52  & 41.41  &45.68& 40.43& 46.26& 45.14& {44.24}\\
Ovanet& 60.30& 51.69& 58.43& 44.61& 70.64& 63.12 & {58.13}\\
Dance& 73.75& 59.42& 78.06& 59.08& 77.63& 67.05 &{69.17}\\
\textbf{UASA}& \textbf{80.82}& \textbf{64.53}& \textbf{83.58}& \textbf{63.15}& \textbf{81.26}& \textbf{72.84} & {\textbf{74.36}} \\ \hline
      \end{tabular}
  }}
  \label{result_opda}
  \end{center}
\end{table*}

\noindent \textbf{Evaluation metrics.}
 We follow \citep{ros} to adopt the \textbf{HOS} score for evaluation. \textbf{HOS} is defined as the harmonic mean between $\mathbf{OS}^*$ and \textbf{UNK} metrics: 
 \begin{equation}
  HOS = 2 \times \mathbf{OS}^* \times \mathbf{UNK}/(\mathbf{OS}^* + \mathbf{UNK}),
\end{equation} 
where $\mathbf{OS}^*$ denotes the average accuracy over the ID classes only and \textbf{UNK} is the recall metric that denotes the ratio of the number of correctly predicted OOD samples over the total number of OOD samples in the target dataset. Since \textbf{HOS} score considers  ID classification and OOD detection simultaneously, we use it as the evaluation metric for CCOD; it is  the most  important metric in our  CCOD task. { To obtain a statistical analysis of the experimental results, we also report the average HOS score of different transfer tasks.}

\noindent \textbf{Compared methods.}
For better reproducibility, we re-implement several state-of-the-art open-source methods: MSP \citep{hendrycks2017baseline}, 
ODIN \citep{liang2018enhancing}, 
OSBP \citep{osbp},
STA \citep{sta},  
ROS \citep{ros}, 
UAN \citep{unida}, 
CMU \citep{cmu}, 
DANCE \citep{dance},   
Energy \citep{liu2020energy}, 
Ovanet \citep{ova}, 
and CIDER \citep{ming2023exploit}.
Since Energy \citep{liu2020energy} is easy to implement and performs well, we treat it as a baseline for the  CCOD task.
Based on the official code and settings, we implement all the methods on all three datasets.

\subsection{Class-Imbalanced Cross-Domain OOD Detection}
As shown in Tables \ref{result_dn}-\ref{result_visda}, our proposed UASA outperforms all compared methods with a large margin for each transfer task, showing the effectiveness of UASA. 
Particularly, on the DomainNet dataset, UASA beats the compared methods by 9.06\% for the P$\rightarrow$R task in Table \ref{result_dn}. The main reason is that the DomainNet dataset contains many classes, and there is little difference between different classes. These small differences make it difficult for the previous methods to distinguish ID and OOD samples. However, our UASA can generate adaptive thresholds for OOD detection to handle small differences between ID and OOD samples. For the Office-Home dataset,  UASA improves the performance by 4.27\% for the RE$\rightarrow$PR task in Table \ref{result_oh} because the types of different domains on the Office-Home dataset vary significantly. The main challenge is aligning various domains. Our proposed UASA can effectively build label-driven prototypes in the source domain, and then conduct domain alignment based on prototypes. As for the VisDA-C dataset, UASA achieves 3.90\% improvement when $\mu=100, C=9$ in Table \ref{result_visda}. The significant improvement is because our UASA can cluster semantically similar samples to eliminate the negative impact of imbalanced classes.

\subsection{Class-Balanced Cross-Domain OOD Detection}

To comprehensively analyze  our model's performance, we compare UASA with some representative state-of-the-art methods under the class-balanced setting. Table \ref{result_opda} reports the corresponding  results. UASA still obtains better performance than these compared methods for all the domain transfer tasks, especially in the RE$\rightarrow$PR task, where UASA outperforms compared methods by 7.07\%. Since the class-balanced setting is easier than the class-imbalanced setting, all methods perform better. UASA still achieves the best performance due to the label-driven prototype and adaptive threshold that handle the domain gap and semantic gap.

\begin{figure}[t!]
\centering
\includegraphics[width=\textwidth]{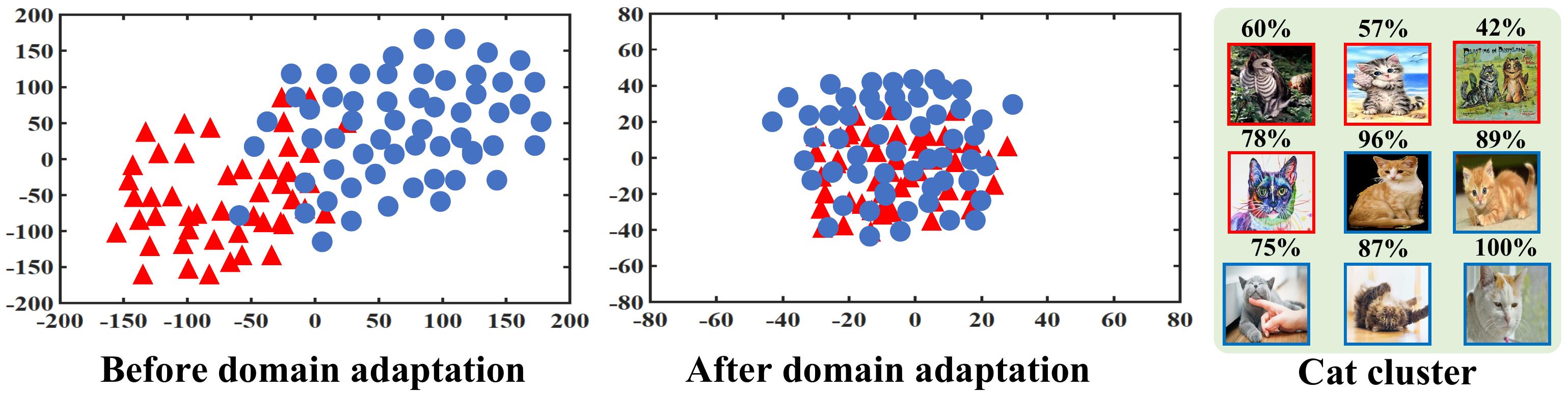}
\caption{Visualizations for  the P$\rightarrow$R task on DomainNet. Left and Middle: T-SNE visualizations of ``before domain adaptation'' (Left) and ``after domain adaptation'' (Middle), where 
\textcolor{red}{red} triangles  denote source  ``cat'' samples and \textcolor{blue}{blue} circles denote target ``cat'' samples. Right: Visualization of our clustering results. We show partial samples from the ``Cat'' cluster, where the labeled percentage is larger than 85\%. We report the probability above each image.
Images with \textcolor{red}{red} edges are from the painting domain. Images with \textcolor{blue}{blue} edges are from the real-world domain.
}
\label{vis}
\end{figure}

\begin{figure}[t!]
\centering
\includegraphics[width=0.32\textwidth]{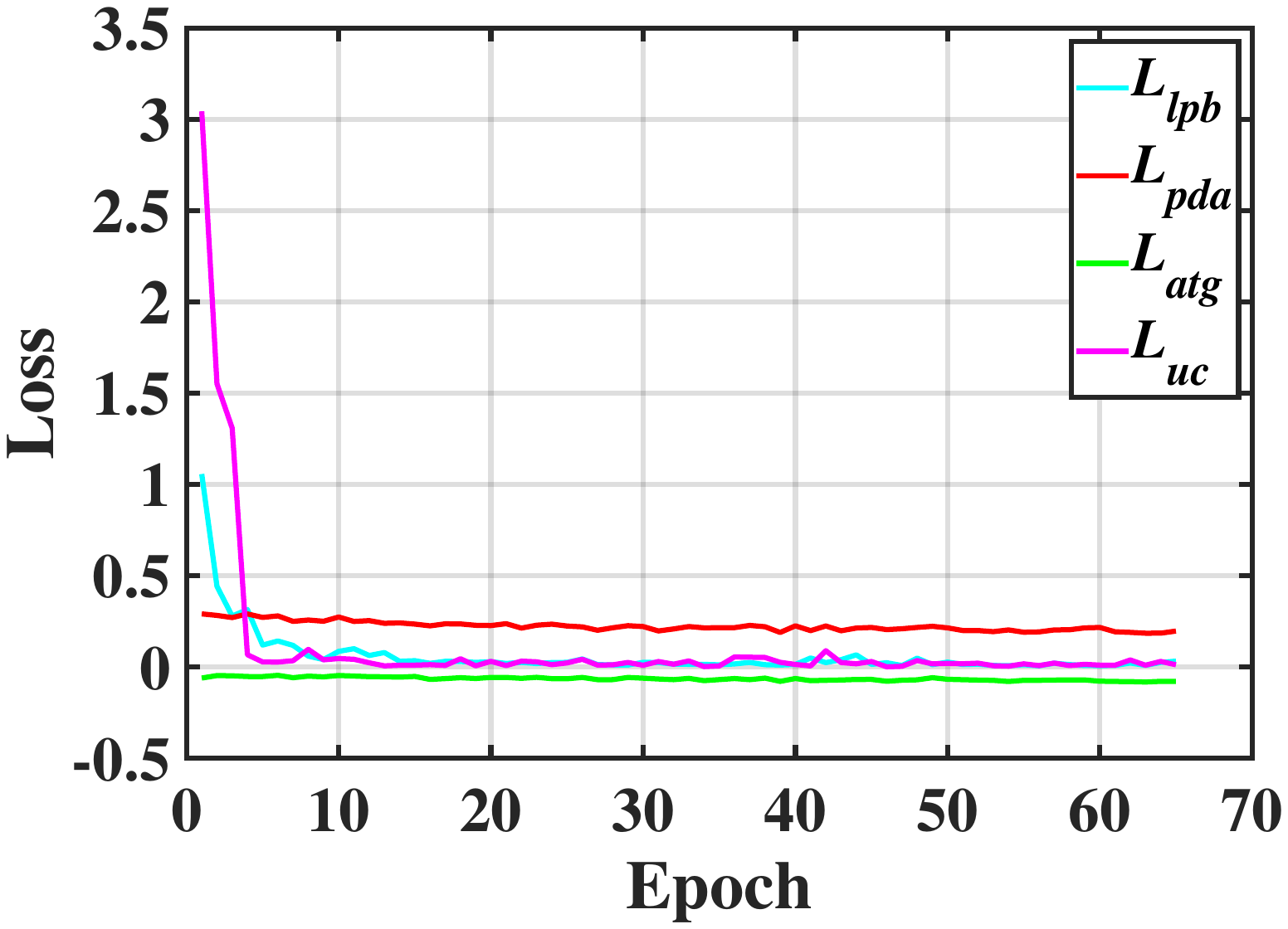} 
\hspace{-0.07in}
\includegraphics[width=0.32\textwidth]{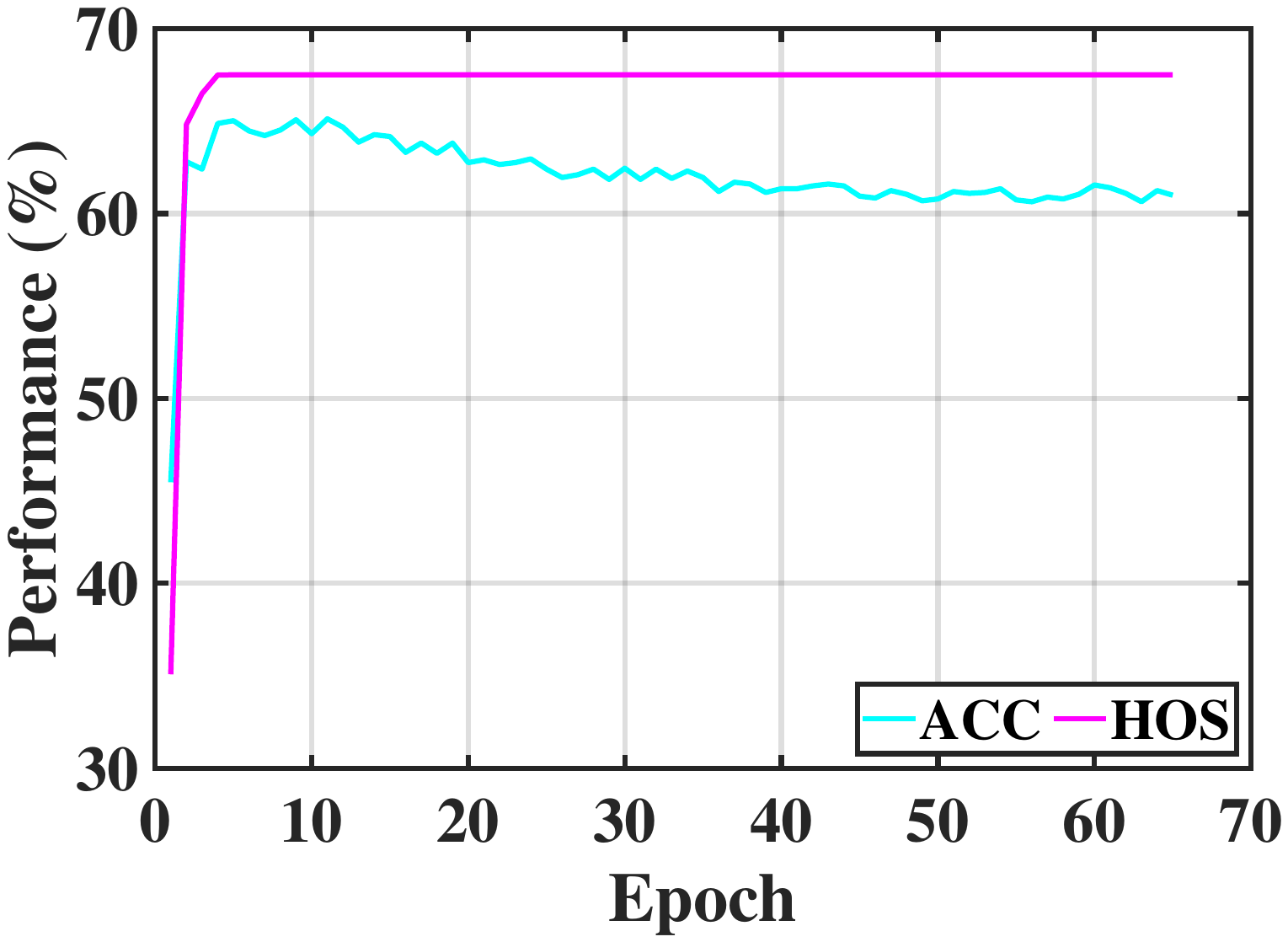} 
\hspace{-0.07in}
\includegraphics[width=0.32\textwidth]{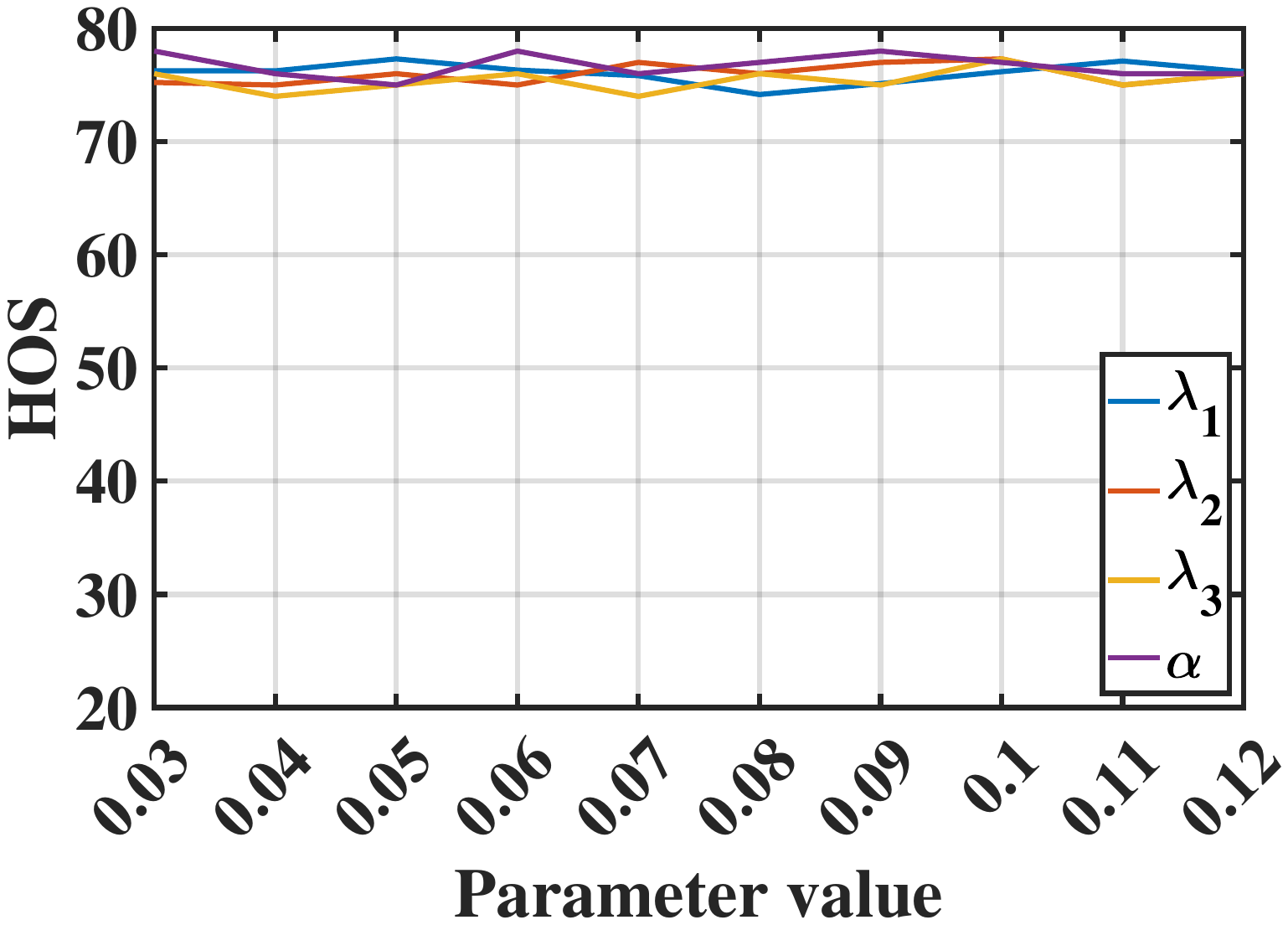}
\caption{Left and Middle: ablative CCOD performance in terms of loss (Left) and performance (Middle) for the P$\rightarrow$R task on the DomainNet dataset across training epochs, where ``ACC'' means ``source classification accuracy''. Right: Parameter sensitivity ($\lambda_1, \lambda_2,\lambda_3,\alpha$).}
\label{fig:xunliantu}
\end{figure}

\noindent \textbf{Feature visualization.}
We randomly choose 40 ``cat'' source samples and 60 target samples to investigate the feature domain distributions in each domain for the P$\rightarrow$R task on DomainNet.
Figure \ref{vis} shows the t-SNE visualizations \citep{van2008visualizing} of ``before domain adaptation''  and ``after domain adaptation'' in our  UASA. Obviously, there is a large distribution gap between the source and target domains, and  UASA can effectively close the domain gap with label-driven prototypes.

\noindent \textbf{Qualitative analysis.}
To qualitatively investigate the effectiveness of UASA,  we report some representative examples on  DomainNet in Figure \ref{vis}.
The visualization results show that  UASA can close the domain gap with high confidence. 

\noindent \textbf{Training process.} We analyze the training process and  performance  in Figure \ref{fig:xunliantu}. We obtain some representative observations:
(i) During training, as the epoch number increases, all the losses ($\mathcal{L}_{lpb}$, $\mathcal{L}_{pda}$, $\mathcal{L}_{atg}$ and $\mathcal{L}_{uc}$) decrease, and the performance (ACC and HOS) increases, illustrating that UASA can both classify source samples and detect target OOD samples simultaneously.
{ (ii) Our full UASA converges quickly and with satisfactory performance within 50 epochs. The main reason is that each designed model mostly contains easy computations with low complexity, \textit{e.g.}, the index operation in Equation \eqref{l_ce}. In Section \ref{source_classification}, we utilize the class weight $m_j$ as the source prototype of the $j$-th class, which also reduces computational cost.
Therefore, our full model has low computational complexity and is more efficient with respect to execution time on the challenging CCOD task.}

{\noindent\textbf{Parameter sensitivity}}. Additionally, we investigate the robustness of the proposed model to different hyper-parameters ($\lambda_1, \lambda_2,\lambda_3,\alpha$) for the P$\rightarrow$R task on the  DomainNet dataset. As shown in Figure \ref{fig:xunliantu},  within a wide range of these hyper-parameters the performance only varies slightly, indicating robustness to different choices of these  parameters. For convenience, we choose $\lambda_1=0.05,\lambda_2=0.1,\lambda_3=0.1$, and $\alpha=0.15$.

\subsection{Ablation Study}
\begin{table}[t!]
  \begin{center}
  \caption{Main ablation study on the VisDA-C dataset, where ``PDA'' means ``prototype-guided domain alignment module'', ``UC'' means ``uncertainty-aware target clustering module'', and ``ATG'' means ``adaptive threshold generation module''.}
  \scalebox{1}{   \setlength{\tabcolsep}{0.5mm}{
    \begin{tabular}{l|ccccccccc}
      \hline
\multirow{2}{*}{Models} & {$\mu=100,$}                                              & {$\mu=50,$}                                               & {$\mu=10,$}    & {$\mu=5,$}                                            & {$\mu=100,$}                                             & {$\mu=50,$}  & {$\mu=10,$} & {$\mu=5,$}   \\
~& $K^s=9$& $K^s=9$& $K^s=9$ & $K^s=9$& $K^s=10$& $K^s=10$& $K^s=10$& $K^s=10$\\\hline
      w/o PDA        & 47.56& 48.13&48.25&42.39 &48.56 &46.03        & 45.33  & 45.86                   \\
      w/o UC & 48.25          & 50.22         & 48.53  & 49.32              & 42.94          & 46.87 & 46.72 & 47.40            \\
             w/o ATG & 48.32          & 49.40          & 49.58 & 50.34 & 41.26          & 48.24 & 47.01   & 47.69                      \\\hline
      {\textbf{Full}}        & \textbf{54.11}& \textbf{54.96}& \textbf{54.07}& \textbf{54.29} &\textbf{48.03}& \textbf{52.16} & \textbf{50.13}& \textbf{51.34} \\ \hline
      \end{tabular}
  }}
  \label{ablation}
  \end{center}
\end{table}

\begin{table}[t!]
    \caption{Effect of the prototype-guided domain alignment module on  the VisDA-C dataset.}
      \begin{center}
    \setlength{\tabcolsep}{0.5mm}{
    \begin{tabular}{l|cccccccc}
    \hline
  \multirow{2}*{Variant} & {$\mu=100,$}                                              & {$\mu=50,$}                                               & {$\mu=10,$}    & {$\mu=5,$}                                            & {$\mu=100,$}                                             & {$\mu=50,$}  & {$\mu=10,$} & {$\mu=5,$}   \\
~& $K^s=9$& $K^s=9$& $K^s=9$ & $K^s=9$& $K^s=10$& $K^s=10$& $K^s=10$& $K^s=10$\\\hline
   KL loss& 52.45& 50.72&53.48& 54.05& \textbf{48.10} & 51.59& \textbf{50.42}& 50.87  \\
   \textbf{Ours}& \textbf{54.11}& \textbf{54.96}& \textbf{54.07}&\textbf{54.29}&  48.03 & \textbf{52.16} & 50.13& \textbf{51.34}\\\hline
$\sigma=0.8$ & 53.55& 54.71& 53.20 & 54.18 & 47.85& 51.76& 49.88& 51.03 \\
$\sigma=0.9$& 53.72& 54.80& 53.88& 54.04& 47.59& 51.82& 50.02& 51.19 \\
$\sigma=1.0$ & \textbf{54.11}& \textbf{54.96}& \textbf{54.07}& \textbf{54.29} &\textbf{48.03}& \textbf{52.16} & \textbf{50.13}& \textbf{51.34}\\
$\sigma=1.1$& 53.94& 54.81& 53.60& 54.15& 47.59& 51.80& 49.86& 51.17 \\
$\sigma=1.2$ & 53.80& 54.66& 53.29& 54.21& 47.84& 51.38& 49.74& 50.92 \\ 
    \hline
    \end{tabular}}
      \end{center}
    \label{tab:ablation^pda}
\end{table}

\noindent \textbf{Main ablation study.}
To evaluate the effectiveness of each module in our UASA, we conduct the main ablation study  on the VisDA-C dataset.
The  corresponding experimental results are in  Table \ref{ablation}. We find that each module significantly contributes to the final performance,  illustrating that the three modules can effectively reduce all three gaps in the CCOD setting (semantic gap, domain gap and class-imbalance). Moreover, the prototype-guided distribution alignment module achieves the largest improvement, demonstrating that it can effectively transfer these label-driven  prototypes from the labeled source domain to these unlabeled target domains for OOD detection. In addition, the distribution alignment module bridges the distribution gap between source and target domains.

\noindent \textbf{Analysis of the prototype-guided domain alignment  module.} To evaluate the ability to close the domain gap, we conduct an ablation study on the prototype-guided domain alignment  module in Table \ref{tab:ablation^pda}, where ``KL loss'' means that we replace the cross-entropy loss in Table \ref{l_nc} with KL loss. Our cross-entropy loss outperforms KL loss in most cases. Also, we test different values of $\sigma$ and achieve the best performance when $\sigma=1.0$.

\begin{table}[t!]
    \caption{Effect of the adaptive threshold generation module on  the VisDA-C dataset.}
      \begin{center}
    \setlength{\tabcolsep}{0.5mm}{
    \begin{tabular}{l|ccccccccccc}
    \hline
  \multirow{2}*{Variant} & {$\mu=100,$}                                              & {$\mu=50,$}                                               & {$\mu=10,$}    & {$\mu=5,$}                                            & {$\mu=100,$}                                             & {$\mu=50,$}  & {$\mu=10,$} & {$\mu=5,$}   \\
~& $K^s=9$& $K^s=9$& $K^s=9$ & $K^s=9$& $K^s=10$& $K^s=10$& $K^s=10$& $K^s=10$\\\hline
 Fixed  & 51.32&50.87&51.69& 52.14& 47.21 & 51.08 & 48.75& 49.72\\
 \textbf{Adaptive} & \textbf{54.11}& \textbf{54.96}& \textbf{54.07}& \textbf{54.29} &\textbf{48.03}& \textbf{52.16} & \textbf{50.13}& \textbf{51.34}\\ 
    \hline
$\Delta=0.3$ & 53.38& 53.75& 53.50& 53.91& 47.95& 51.87& 49.85& 50.40\\
$\Delta=0.4$ & 53.62&54.13&53.76 & 53.92& 47.81& 52.04& 50.04 &50.83 \\
$\Delta=0.5$ & \textbf{54.11}& \textbf{54.96}& \textbf{54.07}& \textbf{54.29} &\textbf{48.03}& \textbf{52.16} & \textbf{50.13}& \textbf{51.34}\\
$\Delta=0.6$ & 54.07& 52.19&53.88& 54.02& 47.92& 51.98& 49.85& 50.27 \\ 
$\Delta=0.7$& 53.75& 53.68& 53.59& 53.77& 47.94& 51.83& 49.70& 50.03\\
    \hline
    \end{tabular}}
      \end{center}
    \label{tab:ablation^threshold}
\end{table}

\begin{table}[t!]
    \caption{Effect of the uncertainty-aware target clustering  module on the VisDA-C dataset.}
      \begin{center}
    \setlength{\tabcolsep}{0.5mm}{
    \begin{tabular}{c|cccccccc}
    \hline
  \multirow{2}*{Variant} & {$\mu=100,$}                                              & {$\mu=50,$}                                               & {$\mu=10,$}    & {$\mu=5,$}                                            & {$\mu=100,$}                                             & {$\mu=50,$}  & {$\mu=10,$} & {$\mu=5,$}   \\
~& $K^s=9$& $K^s=9$& $K^s=9$ & $K^s=9$& $K^s=10$& $K^s=10$& $K^s=10$& $K^s=10$\\\hline
   CE loss& 53.24& 52.19& 53.88& 54
   06& 47.27& 51.89 & 48.50& 49.11 \\
   \textbf{Ours}& \textbf{54.11}& \textbf{54.96}& \textbf{54.07}& \textbf{54.29} &\textbf{48.03}& \textbf{52.16} & \textbf{50.13}& \textbf{51.34}\\\hline
$A=2.4K^s$& 53.85& 54.22& 53.64& 53.97& 47.53& 51.45& 49.83& 50.90 \\
$A=2.4K^s$ & 53.98& 54.15& 53.26& 54.08& 47.82& 51.60 & 49.92& 50.80 \\
$A=2.5K^s$ & \textbf{54.11}& \textbf{54.96}& \textbf{54.07}& \textbf{54.29} &{48.03}& \textbf{52.16} & \textbf{50.13}& \textbf{51.34}\\
$A=2.6K^s$ & 53.48& 53.95& 53.81& 53.97& \textbf{48.21}& 51.84 & 49.88& 51.10 \\ 
$A=2.7K^s$& 53.30& 53.88& 54.01& 53.65& 47.35 & 51.66& 49.75& 50.88\\
    \hline
    \end{tabular}}
      \end{center}
    \label{tab:ablation^uc}
\end{table}

\noindent \textbf{Effect of adaptive threshold generation module.} The adaptive threshold generation module is the core component that directly affects the OOD detection results. To investigate its effectiveness, we implement different variants of the adaptive threshold generation module in Table \ref{tab:ablation^threshold}, where ``Fixed threshold'' indicates that we set each threshold $q_i$ to a constant (we set $q_i={\log(K^s)}/2$) and ``Adaptive threshold'' refers to our full adaptive threshold generation module. In the $\mathcal{L}_{atg}({{p}}_i)$, we introduce the parameter $\Delta$. Table \ref{tab:ablation^threshold} explores a range of values for $\Delta$ with the best performance when $\Delta=0.5$. 

\noindent \textbf{Influence of the uncertainty-aware clustering  module.} We also analyze the influence of our uncertainty-aware clustering  module in Table \ref{tab:ablation^uc}. For  Equation \eqref{l_kl}, we compare the KL loss and cross-entropy loss. We find that the KL loss addresses the class-imbalance better than the cross-entropy loss. In addition, we test different cluster numbers to find the best cluster number $A$. From  Table \ref{tab:ablation^uc}, we can achieve the best performance when $A=2.5K^s$.

\begin{table}[t!]
    \caption{ {Effect of different losses on the VisDA-C dataset.}}
      \begin{center}
    \setlength{\tabcolsep}{0.5mm}{
    \begin{tabular}{c|cccccccc}
    \hline
  {\multirow{2}*{Variant}} & {{$\mu=100,$}}                                              & {{$\mu=50,$} }                                              & {{$\mu=10,$} }   & {{$\mu=5,$} }                                           & {{$\mu=100,$} }   & {{$\mu=50,$}}  & {{$\mu=10,$} } & {{$\mu=5,$} }    \\
~& { $K^s=9$} & { $K^s=9$}& { $K^s=9$} & { $K^s=9$} & { $K^s=10$} & { $K^s=10$}& { $K^s=10$} & { $K^s=10$}  \\\hline
{ w/o $\mathcal{L}_{lpb}$}  & { 51.08} & {52.89} & {51.90}& { 52.37}& {47.82}& {51.88}& {48.65}& {50.10} \\
{w/o $\mathcal{L}_{pda}$} & {51.85}& {52.76}& {52.44}& {51.98}& {46.23}& {51.50}& {48.72}& {49.59} \\
{w/o $\mathcal{L}_{atg}$} & { 52.10}& {53.08}& {52.95}& {52.87}& {47.18}& {50.80}& {49.24}&{50.28} \\
{w/o $\mathcal{L}_{uc}$} & { 52.36}& {53.67}& {53.15}& {53.28}& {47.70}& {51.22}& {49.08}& {50.84}\\\hline
  { \textbf{Full} }& {\textbf{54.11}}& {\textbf{54.96}}& {\textbf{54.07}}& {\textbf{54.29}} &{\textbf{48.03}}& {\textbf{52.16}} & {\textbf{50.13}} & {\textbf{51.34}}\\\hline
    \end{tabular}}
      \end{center}
    \label{tab:ablation_loss}
\end{table}

{ \noindent \textbf{Impact of different loss functions.} In our proposed UASA, we introduce different losses to supervise different modules.  We remove individual losses in Table~\ref{tab:ablation_loss} and  observe that we achieve the best performance when we utilize all the losses for model training. It demonstrates that each loss is effective in our proposed UASA during  training.}

\section{Conclusion}
\label{con}
In this paper, we pose a novel yet challenging  setting for an OOD detection system:  CCOD,  where we  consider three gaps: the semantic gap between ID and OOD classes, the domain gap between source and target domains, and the class-imbalance between different classes. To tackle this challenging setting, we propose a novel uncertainty-aware adaptive semantic alignment (UASA) network to handle all three gaps.  
Experimental results on three challenging datasets show the effectiveness of  UASA. 
In representative cases,  UASA outperforms all compared methods by 9.06\% on the DomainNet dataset.
{ In the future, we will extend our proposed UASA system to more complex video datasets to further improve its
generalization ability. We also aim to explore different ways to align
source/target domains to improve the model. }

\section{Acknowledgement}
\label{ack}
This research is part of the programme DesCartes and is supported by the National Research Foundation, Prime Minister’s Office, Singapore under its Campus for Research Excellence and Technological Enterprise (CREATE) programme.
{ We also acknowledge the native speaker, Michael Yuhas John (Email: michaelj004@e.ntu.edu.sg), for helping us  check and polish our paper.}
\bibliographystyle{elsarticle-num} 
 \bibliography{paper}

@inproceedings{resnet,
  title={Deep residual learning for image recognition},
  author={He, Kaiming and Zhang, Xiangyu and Ren, Shaoqing and Sun, Jian},
  booktitle={Proceedings of the IEEE conference on computer vision and pattern recognition},
  pages={770--778},
  year={2016}
}

@article{fang2023hierarchical,
  title={Hierarchical local-global transformer for temporal sentence grounding},
  author={Fang, Xiang and Liu, Daizong and Zhou, Pan and Xu, Zichuan and Li, Ruixuan},
  journal={IEEE Transactions on Multimedia},
  year={2023},
  publisher={IEEE}
}

@article{fang2022multi,
  title={Multi-modal cross-domain alignment network for video moment retrieval},
  author={Fang, Xiang and Liu, Daizong and Zhou, Pan and Hu, Yuchong},
  journal={IEEE Transactions on Multimedia},
  volume={25},
  pages={7517--7532},
  year={2022},
  publisher={IEEE}
}

@inproceedings{fang2026cogniVerse,
  title={CogniVerse: Revolutionizing Multi-modal Retrieval-Augmented Generation with Cognitive Reflection and Geometric Reasoning},
  author={Fang, Xiang and Fang, Wanlong and Wang, Changshuo},
  booktitle={Proceedings of the IEEE/CVF Conference on Computer Vision and Pattern Recognition},
  year={2026}
}

@inproceedings{fang2023you,
  title={You can ground earlier than see: An effective and efficient pipeline for temporal sentence grounding in compressed videos},
  author={Fang, Xiang and Liu, Daizong and Zhou, Pan and Nan, Guoshun},
  booktitle={Proceedings of the IEEE/CVF Conference on Computer Vision and Pattern Recognition},
  pages={2448--2460},
  year={2023}
}

@inproceedings{fang2025hierarchical,
  title={Hierarchical Semantic-Augmented Navigation: Optimal Transport and Graph-Driven Reasoning for Vision-Language Navigation},
  author={Fang, Xiang and Fang, Wanlong and Wang, Changshuo},
  booktitle={Advances in Neural Information Processing Systems},
  year={2025}
}

@inproceedings{fang2025adaptive,
  title={Adaptive Multi-prompt Contrastive Network for Few-shot Out-of-distribution Detection},
  author={Fang, Xiang and Easwaran, Arvind and Genest, Blaise},
  booktitle={International Conference on Machine Learning},
  year={2025}
}

@inproceedings{fang2026slap,
  title={SLAP: The Semantic Least Action Principle for Variational Video-Language Modeling},
  author={Fang, Xiang and Fang, Wanlong},
  booktitle={International Conference on Machine Learning},
  year={2026}
}

@inproceedings{fang2026immuno,
  title={Immuno-VLM: Immunizing Large Vision-Language Models via Generative Semantic Antibodies for Open-World Trustworthiness},
  author={Fang, Xiang and Fang, Wanlong and Ji, Wei},
  booktitle={International Conference on Machine Learning},
  year={2026}
}

@inproceedings{fang2026disentangling,
  title={Disentangling Adversarial Prompts: A Semantic-Graph Defense for Robust LLM Security},
  author={Fang, Xiang and Fang, Wanlong},
 booktitle={Proceedings of the AAAI Conference on Artificial Intelligence},
year={2026}
}

@inproceedings{fang2026advancing,
  title={Advancing Out-of-Distribution Detection Across Diverse Scenarios},
  author={Fang, Xiang},
  booktitle={Proceedings of the AAAI Conference on Artificial Intelligence},
  volume={40},
  number={48},
  pages={41042--41043},
  year={2026}
}

@inproceedings{fang2026unveiling,
  title={Unveiling the Fragility of Vision-Language Models: Multi-Modal Adversarial Synergy via Texture-Constrained Perturbations and Cross-Modal Optimization},
  author={Fang, Xiang and Fang, Wanlong and Wang, Changshuo},
 booktitle={Proceedings of the AAAI Conference on Artificial Intelligence},
year={2026}
}

@inproceedings{fang2026rethinking,
  title={Rethinking Video-language Model From the Language Input Perspective},
  author={Fang, Xiang and Fang, Wanlong and Wang, Changshuo and Qu, Xiaoye and Liu, Daizong},
 booktitle={Proceedings of the AAAI Conference on Artificial Intelligence},
year={2026}
}

@inproceedings{fang2026towards,
  title={Towards Unified Vision-Language Models With Incomplete Multi-Modal Inputs},
  author={Fang, Xiang and Fang, Wanlong and Wang, Changshuo and Tang, Keke and Liu, Daizong and Wang, Siyi and Ji, Wei},
 booktitle={Proceedings of the AAAI Conference on Artificial Intelligence},
year={2026}
}

@inproceedings{fang2025multi,
  title={Multi-pair temporal sentence grounding via multi-thread knowledge transfer network},
  author={Fang, Xiang and Fang, Wanlong and Wang, Changshuo and Liu, Daizong and Tang, Keke and Dong, Jianfeng and Zhou, Pan and Li, Beibei},
  booktitle={Proceedings of the AAAI Conference on Artificial Intelligence},
  volume={39},
  number={3},
  pages={2915--2923},
  year={2025}
}

@inproceedings{fang2024fewer,
  title={Fewer Steps, Better Performance: Efficient Cross-Modal Clip Trimming for Video Moment Retrieval Using Language},
  author={Fang, Xiang and Liu, Daizong and Fang, Wanlong and Zhou, Pan and Xu, Zichuan and Xu, Wenzheng and Chen, Junyang and Li, Renfu},
  booktitle={Proceedings of the AAAI Conference on Artificial Intelligence},
  volume={38},
  number={2},
  pages={1735--1743},
  year={2024}
}

@inproceedings{fang2024multi,
  title={Multi-Pair Temporal Sentence Grounding via Multi-Thread Knowledge Transfer Network},
  author={Fang, Xiang and Fang, Wanlong and Wang, Changshuo and Liu, Daizong and Tang, Keke and Dong, Jianfeng and Zhou, Pan and Li, Beibei},
  booktitle={Proceedings of the AAAI Conference on Artificial Intelligence},
  year={2025}
}

@inproceedings{fang2025turing,
  title={Turing Patterns for Multimedia: Reaction-Diffusion Multi-Modal Fusion for Language-Guided Video Moment Retrieval},
  author={Fang, Xiang and Fang, Wanlong and Ji, Wei and Chua, Tat-Seng},
  booktitle={ACM International Conference on Multimedia},
  year={2025}
}

@inproceedings{fang2024not,
  title={Not all inputs are valid: Towards open-set video moment retrieval using language},
  author={Fang, Xiang and Fang, Wanlong and Liu, Daizong and Qu, Xiaoye and Dong, Jianfeng and Zhou, Pan and Li, Renfu and Xu, Zichuan and Chen, Lixing and Zheng, Panpan and others},
  booktitle={Proceedings of the 32nd ACM International Conference on Multimedia},
  pages={28--37},
  year={2024}
}

@inproceedings{fang2024rethinking,
  title={Rethinking Weakly-supervised Video Temporal Grounding From a Game Perspective},
  author={Fang, Xiang and Xiong, Zeyu and Fang, Wanlong and Qu, Xiaoye and Chen, Chen and Dong, Jianfeng and Tang, Keke and Zhou, Pan and Cheng, Yu and Liu, Daizong},
  booktitle={European Conference on Computer Vision},
  year={2024},
  organization={Springer}
}

@inproceedings{fang2023annotations,
  title={Annotations Are Not All You Need: A Cross-modal Knowledge Transfer Network for Unsupervised Temporal Sentence Grounding},
  author={Fang, Xiang and Liu, Daizong and Fang, Wanlong and Zhou, Pan and Cheng, Yu and Tang, Keke and Zou, Kai},
  booktitle={Findings of the Association for Computational Linguistics: EMNLP 2023},
  pages={8721--8733},
  year={2023}
}

@article{fang2021unbalanced,
  title={Unbalanced incomplete multi-view clustering via the scheme of view evolution: Weak views are meat; strong views do eat},
  author={Fang, Xiang and Hu, Yuchong and Zhou, Pan and Wu, Dapeng Oliver},
  journal={IEEE Transactions on Emerging Topics in Computational Intelligence},
  volume={6},
  number={4},
  pages={913--927},
  year={2021},
  publisher={IEEE}
}

@article{fang2025adaptivetai,
  title={Adaptive Hierarchical Graph Cut for Multi-granularity Out-of-distribution Detection},
  author={Fang, Xiang and Easwaran, Arvind and Genest, Blaise and Suganthan, Ponnuthurai Nagaratnam},
  journal={IEEE Transactions on Artificial Intelligence},
  year={2025}
}

@article{fang2021animc,
  title={Animc: A soft approach for autoweighted noisy and incomplete multiview clustering},
  author={Fang, Xiang and Hu, Yuchong and Zhou, Pan and Wu, Dapeng},
  journal={IEEE Transactions on Artificial Intelligence},
  volume={3},
  number={2},
  pages={192--206},
  year={2021},
  publisher={IEEE}
}

@article{fang2020v,
  title={V3H: View variation and view heredity for incomplete multiview clustering},
  author={Fang, Xiang and Hu, Yuchong and Zhou, Pan and Wu, Dapeng Oliver},
  journal={IEEE Transactions on Artificial Intelligence},
  volume={1},
  number={3},
  pages={233--247},
  year={2020},
  publisher={IEEE}
}

@article{fang2020double,
  title={Double self-weighted multi-view clustering via adaptive view fusion},
  author={Fang, Xiang and Hu, Yuchong},
  journal={arXiv preprint arXiv:2011.10396},
  year={2020}
}

@article{liu2023exploring,
  title={Exploring optical-flow-guided motion and detection-based appearance for temporal sentence grounding},
  author={Liu, Daizong and Fang, Xiang and Hu, Wei and Zhou, Pan},
  journal={IEEE Transactions on Multimedia},
  volume={25},
  pages={8539--8553},
  year={2023},
  publisher={IEEE}
}

@inproceedings{wang2025taylor,
  title={Taylor series-inspired local structure fitting network for few-shot point cloud semantic segmentation},
  author={Wang, Changshuo and He, Shuting and Fang, Xiang and Wu, Meiqing and Lam, Siew-Kei and Tiwari, Prayag},
  booktitle={Proceedings of the AAAI Conference on Artificial Intelligence},
  volume={39},
  number={7},
  pages={7527--7535},
  year={2025}
}

@inproceedings{wang2025point,
  title={Point clouds meets physics: Dynamic acoustic field fitting network for point cloud understanding},
  author={Wang, Changshuo and He, Shuting and Fang, Xiang and Han, Jiawei and Liu, Zhonghang and Ning, Xin and Li, Weijun and Tiwari, Prayag},
  booktitle={Proceedings of the Computer Vision and Pattern Recognition Conference},
  pages={22182--22192},
  year={2025}
}

@inproceedings{wang2025dypolyseg,
  title={DyPolySeg: Taylor Series-Inspired Dynamic Polynomial Fitting Network for Few-shot Point Cloud Semantic Segmentation},
  author={Wang, Changshuo and Fang, Xiang and Tiwari, Prayag},
  booktitle={Forty-second International Conference on Machine Learning},
  year={2025}
}

@article{wang2026reasoning,
  title={Reasoning beyond points: A visual introspective approach for few-shot 3d segmentation},
  author={Wang, Changshuo and He, Shuting and Fang, Xiang and Hu, Zhijian and Huang, Jia-Hong and Shen, Yixian and Tiwari, Prayag},
  journal={Advances in Neural Information Processing Systems},
  volume={38},
  pages={117394--117414},
  year={2026}
}

@article{wang2026from,
  title={From Coarse to Fine: Deep Prototype Refinement Network for Few-Shot Point Cloud Semantic Segmentation},
  author={Wang, Changshuo and He, Shuting and Fang, Xiang and Li, Weijun and Gao, Xingyu and Liu, Zhonghang and Tiwari, Prayag and Kanoulas, Dimitrios},
  journal={International Conference on Machine Learning},
  year={2026}
}

@article{wang2026topadapter,
  title={TopAdapter: Topology-Aware Prompt Tuning for Efficient Point Cloud Understanding},
  author={Wang, Changshuo and He, Shuting and Fang, Xiang and Li, Weijun and Shen, Yixian and Xu, Mingkun and Sun, Zhongtian and Tiwari, Prayag},
  journal={International Conference on Machine Learning},
  year={2026}
}

@inproceedings{wang2026biologically,
  title={Biologically-Inspired Evolutionary Domain Symbiosis for Few-shot and Zero-shot Point Cloud Semantic Segmentation},
  author={Wang, Changshuo and Hu, Zhijian and Fang, Xiang and Yu, Zai Yang and Wu, Yibin and Xu, Mingkun and Wang, Yusong and Gao, Xingyu and Tiwari, Prayag},
  booktitle={Proceedings of the AAAI Conference on Artificial Intelligence},
  volume={40},
  number={12},
  pages={9666--9674},
  year={2026}
}

@inproceedings{yang2025eood,
  title={EOOD: Entropy-based Out-of-distribution Detection},
  author={Yang, Guide and Hou, Chao and Peng, Weilong and Fang, Xiang and Nie, Yongwei and Zhu, Peican and Tang, Keke},
  booktitle={2025 International Joint Conference on Neural Networks (IJCNN)},
  pages={1--8},
  year={2025},
  organization={IEEE}
}

@inproceedings{wang2025reducing
,
  title={Reducing T-Depth and T-Count in Quantum Multiplication Using Compressor Primitives},
  author={Wang, Siyi and Dutta, Suman and Lee, Wei Jie Bryan and Feng, Jerrie and Fang, Xiang and Chattopadhyay, Anupam},
  booktitle={Proceedings of the Great Lakes Symposium on VLSI 2025},
  pages={35--40},
  year={2025}
}

@inproceedings{lei2025exploring,
  title={Exploring Disentangled Appearance-Motion Contexts for Temporal Activity Localization},
  author={Lei, Huashuo and Cai, Xiaowen and Liu, Daizong and Fang, Xiang and Qu, Xiaoye and Dong, Jianfeng and Yu, Jixiang and Jin, Keyan},
  booktitle={2025 International Joint Conference on Neural Networks (IJCNN)},
  pages={1--8},
  year={2025},
  organization={IEEE}
}

@inproceedings{zhang2025monoattack,
  title={MonoAttack: A Strong Attack Framework with Depth-Migration and Attribute-Tampering for Monocular 3D Object Detection},
  author={Zhang, Xiayue and Lei, Huashuo and Liu, Daizong and Qu, Xiaoye and Fang, Xiang and Guan, Runwei and Jin, Keyan},
  booktitle={2025 International Joint Conference on Neural Networks (IJCNN)},
  pages={1--8},
  year={2025},
  organization={IEEE}
}

@inproceedings{zhang2025manipulating,
  title={Manipulating the Bounding Box: Multimodal Controlled Backdoor Attacks on 3D Visual Grounding Models},
  author={Zhang, Xiayue and Lei, Huashuo and Liu, Daizong and Qu, Xiaoye and Fang, Xiang and Guan, Runwei and Jin, Keyan},
  booktitle={2025 International Joint Conference on Neural Networks (IJCNN)},
  pages={1--8},
  year={2025},
  organization={IEEE}
}

@article{wang2025prototype,
  title={Prototype-driven structure synergy network for remote sensing images segmentation},
  author={Wang, Junyi and Li, Jinjiang and Fan, Guodong and Ju, Yakun and Fang, Xiang and Kot, Alex C},
  journal={IEEE Transactions on Geoscience and Remote Sensing},
  year={2025},
  publisher={IEEE}
}

@inproceedings{wang2025seeing,
  title={Seeing the Overlooked: Bio-Visual Inspired Weak Saliency Feedback Transformer for Person Re-identification},
  author={Wang, Changshuo and He, Shuting and Fang, Xiang and Nan, Fangzhe and Tiwari, Prayag},
  booktitle={Proceedings of the 33rd ACM International Conference on Multimedia},
  pages={3192--3201},
  year={2025}
}

@inproceedings{fang2026align,
  title={To align or not to align: Strategic multimodal representation alignment for optimal performance},
  author={Fang, Wanlong and Zhang, Tianle and Chan, Alvin},
  booktitle={Proceedings of the AAAI Conference on Artificial Intelligence},
  volume={40},
  number={25},
  pages={21056--21064},
  year={2026}
}

@article{liu2023conditional,
  title={Conditional video diffusion network for fine-grained temporal sentence grounding},
  author={Liu, Daizong and Zhu, Jiahao and Fang, Xiang and Xiong, Zeyu and Wang, Huan and Li, Renfu and Zhou, Pan},
  journal={IEEE Transactions on Multimedia},
  volume={26},
  pages={5461--5476},
  year={2023},
  publisher={IEEE}
}

@article{liu2024pandora,
  title={Pandora's box: Towards building universal attackers against real-world large vision-language models},
  author={Liu, Daizong and Yang, Mingyu and Qu, Xiaoye and Zhou, Pan and Fang, Xiang and Tang, Keke and Wan, Yao and Sun, Lichao},
  journal={Advances in Neural Information Processing Systems},
  volume={37},
  pages={52127--52158},
  year={2024}
}

@inproceedings{liu2026attacking,
  title={Attacking Gray-Box Large Vision-Language Models with Adaptive SVD-Structured Adversarial Alignment},
  author={Liu, Daizong and Cai, Xiaowen and Dong, Junhao and Guo, Zhongliang and Qu, Xiaoye and Guan, Runwei and Fang, Xiang and Ye, Dengpan},
  booktitle={International Conference on Machine Learning},
  year={2026}
}

@inproceedings{liu2024unsupervised,
  title={Unsupervised domain adaptative temporal sentence localization with mutual information maximization},
  author={Liu, Daizong and Fang, Xiang and Qu, Xiaoye and Dong, Jianfeng and Yan, He and Yang, Yang and Zhou, Pan and Cheng, Yu},
  booktitle={Proceedings of the AAAI Conference on Artificial Intelligence},
  volume={38},
  number={4},
  pages={3567--3575},
  year={2024}
}

@inproceedings{liu2023hypotheses,
  title={Hypotheses tree building for one-shot temporal sentence localization},
  author={Liu, Daizong and Fang, Xiang and Zhou, Pan and Di, Xing and Lu, Weining and Cheng, Yu},
  booktitle={Proceedings of the AAAI Conference on Artificial Intelligence},
  volume={37},
  number={2},
  pages={1640--1648},
  year={2023}
}

@inproceedings{tang2024reparameterization,
  title={Reparameterization head for efficient multi-input networks},
  author={Tang, Keke and Zhao, Wenyu and Peng, Weilong and Fang, Xiang and Cui, Xiaodong and Zhu, Peican and Tian, Zhihong},
  booktitle={ICASSP 2024-2024 IEEE International Conference on Acoustics, Speech and Signal Processing (ICASSP)},
  pages={6190--6194},
  year={2024},
  organization={IEEE}
}

@article{xiong2024rethinking,
  title={Rethinking video sentence grounding from a tracking perspective with memory network and masked attention},
  author={Xiong, Zeyu and Liu, Daizong and Fang, Xiang and Qu, Xiaoye and Dong, Jianfeng and Zhu, Jiahao and Tang, Keke and Zhou, Pan},
  journal={IEEE Transactions on Multimedia},
  volume={26},
  pages={11204--11218},
  year={2024},
  publisher={IEEE}
}

@inproceedings{tang2025simplification,
  title={Simplification is all you need against out-of-distribution overconfidence},
  author={Tang, Keke and Hou, Chao and Peng, Weilong and Fang, Xiang and Wu, Zhize and Nie, Yongwei and Wang, Wenping and Tian, Zhihong},
  booktitle={Proceedings of the Computer Vision and Pattern Recognition Conference},
  pages={5030--5040},
  year={2025}
}

@article{cai2026towards,
  title={Towards building model/prompt-transferable attackers against large vision-language models},
  author={Cai, Xiaowen and Liu, Daizong and Qu, Xiaoye and Fang, Xiang and Dong, Jianfeng and Tang, Keke and Zhou, Pan and Sun, Lichao and Hu, Wei},
  journal={Advances in Neural Information Processing Systems},
  volume={38},
  pages={174022--174058},
  year={2026}
}

@article{yan2026fit,
  title={Fit the distribution: Cross-image/prompt adversarial attacks on multimodal large language models},
  author={Yan, Hai and Ma, Haijian and Cai, Xiaowen and Liu, Daizong and Yuan, Zenghui and Qu, Xiaoye and Dong, Jianfeng and Guan, Runwei and Fang, Xiang and He, Hongyang and others},
  journal={Advances in Neural Information Processing Systems},
  volume={38},
  pages={75204--75247},
  year={2026}
}

@inproceedings{liu2024towards,
  title={Towards robust temporal activity localization learning with noisy labels},
  author={Liu, Daizong and Qu, Xiaoye and Fang, Xiang and Dong, Jianfeng and Zhou, Pan and Nan, Guoshun and Tang, Keke and Fang, Wanlong and Cheng, Yu},
  booktitle={Proceedings of the 2024 Joint International Conference on Computational Linguistics, Language Resources and Evaluation (LREC-COLING 2024)},
  pages={16630--16642},
  year={2024}
}

@inproceedings{cai2025imperceptible,
  title={Imperceptible Beam-Sensitive Adversarial Attacks for LiDAR-based Object Detection in Autonomous Driving},
  author={Cai, Fuyao and Liu, Daizong and Fang, Xiang and Yu, Jixiang and Tang, Keke and Zhou, Pan},
  booktitle={2025 IEEE International Conference on Multimedia and Expo (ICME)},
  pages={1--6},
  year={2025},
  organization={IEEE}
}

@article{kuai2026dynamic,
  title={Dynamic Graph-enhanced Event Refinement for Temporal Sentence Grounding of Micro-moments},
  author={Kuai, Mingjin and Qin, You and Fang, Xiang and Ji, Wei and Zimmermann, Roger},
  journal={IEEE Transactions on Multimedia},
  year={2026},
  publisher={IEEE}
}

@inproceedings{fang2026towardsicml,
  title={Towards Understanding Modality Interaction in Multimodal Language Models via Partial Information Decomposition},
  author={Fang, Wanlong and Zhang, Tianle and Tao, Wen and Chan, Alvin},
  booktitle={International Conference on Machine Learning},
  year={2026}
}

@article{maqsood2024mox,
  title={MOX-NET: Multi-stage deep hybrid feature fusion and selection framework for monkeypox classification},
  author={Maqsood, Sarmad and Dama{\v{s}}evi{\v{c}}ius, Robertas and Shahid, Sana and Forkert, Nils D},
  journal={Expert Systems with Applications},
  volume={255},
  pages={124584},
  year={2024},
  publisher={Elsevier}
}

@article{atmakuru2024deep,
  title={Deep learning in radiology for lung cancer diagnostics: A systematic review of classification, segmentation, and predictive modeling techniques},
  author={Atmakuru, Anirudh and Chakraborty, Subrata and Faust, Oliver and Salvi, Massimo and Barua, Prabal Datta and Molinari, Filippo and Acharya, UR and Homaira, Nusrat},
  journal={Expert Systems with Applications},
  pages={124665},
  year={2024},
  publisher={Elsevier}
}

@article{yao2024open,
  title={Open-set adversarial domain match for electronic nose drift compensation and unknown gas recognition},
  author={Yao, Youbin and Chen, Bin and Liu, Chuanjun and Feng, Changhao and Gao, Xuliang and Gu, Yun},
  journal={Expert Systems with Applications},
  volume={250},
  pages={123757},
  year={2024},
  publisher={Elsevier}
}

@article{ren2024towards,
  title={Towards unsupervised domain adaptation via domain-transformer},
  author={Ren, Chuan-Xian and Zhai, Yiming and Luo, You-Wei and Yan, Hong},
  journal={International Journal of Computer Vision},
  pages={1--21},
  year={2024},
  publisher={Springer}
}

@article{xie2024adapting,
  title={Adapting Across Domains via Target-Oriented Transferable Semantic Augmentation Under Prototype Constraint},
  author={Xie, Mixue and Li, Shuang and Gong, Kaixiong and Wang, Yulin and Huang, Gao},
  journal={International Journal of Computer Vision},
  volume={132},
  number={4},
  pages={1417--1441},
  year={2024},
  publisher={Springer}
}

@article{sun2025prototype,
  title={Prototype-Optimized unsupervised domain adaptation via dynamic Transformer encoder for sensor drift compensation in electronic nose systems},
  author={Sun, Jie and Zheng, Hao and Diao, Wenxing and Sun, Zhilin and Qi, Zhengdong and Wang, Xiaozheng},
  journal={Expert Systems with Applications},
  volume={260},
  pages={125444},
  year={2025},
  publisher={Elsevier}
}

@article{fang2024prototype,
  title={Prototype learning for adversarial domain adaptation},
  author={Fang, Yuchun and Chen, Chen and Zhang, Wei and Wu, Jiahua and Zhang, Zhaoxiang and Xie, Shaorong},
  journal={Pattern Recognition},
  pages={110653},
  year={2024},
  publisher={Elsevier}
}

@inproceedings{cui2024effective,
  title={Effective Comparative Prototype Hashing for Unsupervised Domain Adaptation},
  author={Cui, Hui and Zhao, Lihai and Li, Fengling and Zhu, Lei and Han, Xiaohui and Li, Jingjing},
  booktitle={Proceedings of the AAAI Conference on Artificial Intelligence},
  volume={38},
  number={8},
  pages={8329--8337},
  year={2024}
}

@inproceedings{belal2024multi,
  title={Multi-Source Domain Adaptation for Object Detection with Prototype-based Mean Teacher},
  author={Belal, Atif and Meethal, Akhil and Romero, Francisco Perdigon and Pedersoli, Marco and Granger, Eric},
  booktitle={Proceedings of the IEEE/CVF Winter Conference on Applications of Computer Vision},
  pages={1277--1286},
  year={2024}
}

@article{gao2024learning,
  title={Learning transferable conceptual prototypes for interpretable unsupervised domain adaptation},
  author={Gao, Junyu and Ma, Xinhong and Xu, Changsheng},
  journal={IEEE Transactions on Image Processing},
  year={2024},
  publisher={IEEE}
}

@article{yang2022unsupervised,
  title={Unsupervised domain adaptation for disguised-gait-based person identification on micro-Doppler signatures},
  author={Yang, Yang and Yang, Xiaoyi and Sakamoto, Takuya and Fioranelli, Francesco and Li, Beichen and Lang, Yue},
  journal={IEEE Transactions on Circuits and Systems for Video Technology},
  volume={32},
  number={9},
  pages={6448--6460},
  year={2022},
  publisher={IEEE}
}

@article{ainam2021unsupervised,
  title={Unsupervised domain adaptation for person re-identification with iterative soft clustering},
  author={Ainam, Jean-Paul and Qin, Ke and Owusu, Jim Wilson and Lu, Guoming},
  journal={Knowledge-Based Systems},
  volume={212},
  pages={106644},
  year={2021},
  publisher={Elsevier}
}

@article{zhang2020unsupervised,
  title={Unsupervised multi-class domain adaptation: Theory, algorithms, and practice},
  author={Zhang, Yabin and Deng, Bin and Tang, Hui and Zhang, Lei and Jia, Kui},
  journal={IEEE Transactions on Pattern Analysis and Machine Intelligence},
  volume={44},
  number={5},
  pages={2775--2792},
  year={2020},
  publisher={IEEE}
}

@article{cui2024unified,
  title={Unified bi-encoder bispace-discriminator disentanglement for cross-domain echocardiography segmentation},
  author={Cui, Xiaoxiao and Wang, Boyu and Jiang, Shanzhi and Liu, Zhi and Xu, Hongji and Cui, Lizhen and Li, Shuo},
  journal={Knowledge-Based Systems},
  volume={303},
  pages={112394},
  year={2024},
  publisher={Elsevier}
}

@article{liu2024learning,
  title={Learning With Fewer Labels in Computer Vision},
  author={Liu, Li and Hospedales, Timothy and LeCun, Yann and Long, Mingsheng and Luo, Jiebo and Ouyang, Wanli and Pietik{\"a}inen, Matti and Tuytelaars, Tinne},
  journal={IEEE Transactions on Pattern Analysis and Machine Intelligence},
  volume={46},
  number={3},
  pages={1319--1326},
  year={2024},
  publisher={IEEE}
}

@article{zhang2024cross,
  title={Cross-domain data fusion generation: A novel composite label-guided generative solution for adaptation diagnosis},
  author={Zhang, Tian and Lin, Jing and Jiao, Jinyang and Li, Hao},
  journal={Knowledge-Based Systems},
  volume={301},
  pages={112284},
  year={2024},
  publisher={Elsevier}
}

@article{fang2024source,
  title={Source-free unsupervised domain adaptation: A survey},
  author={Fang, Yuqi and Yap, Pew-Thian and Lin, Weili and Zhu, Hongtu and Liu, Mingxia},
  journal={Neural Networks},
  pages={106230},
  year={2024},
  publisher={Elsevier}
}

@article{ullah2024video,
  title={Video domain adaptation for semantic segmentation using perceptual consistency matching},
  author={Ullah, Ihsan and An, Sion and Kang, Myeongkyun and Chikontwe, Philip and Lee, Hyunki and Choi, Jinwoo and Park, Sang Hyun},
  journal={Neural Networks},
  volume={179},
  pages={106505},
  year={2024},
  publisher={Elsevier}
}

@article{wang2025slbdetection,
  title={SLBDetection-Net: Towards closed-set and open-set student learning behavior detection in smart classroom of K-12 education},
  author={Wang, Zhifeng and Li, Longlong and Zeng, Chunyan and Dong, Shi and Sun, Jianwen},
  journal={Expert Systems with Applications},
  volume={260},
  pages={125392},
  year={2025},
  publisher={Elsevier}
}

@article{li2024auto,
  title={An auto-regulated universal domain adaptation network for uncertain diagnostic scenarios of rotating machinery},
  author={Li, Jipu and Zhang, Xiaoge and Yue, Ke and Chen, Junbin and Chen, Zhuyun and Li, Weihua},
  journal={Expert Systems with Applications},
  volume={249},
  pages={123836},
  year={2024},
  publisher={Elsevier}
}

@article{jiao2024open,
  title={Open-set recognition with long-tail sonar images},
  author={Jiao, Wenpei and Zhang, Jianlei and Zhang, Chunyan},
  journal={Expert Systems with Applications},
  volume={249},
  pages={123495},
  year={2024},
  publisher={Elsevier}
}

@article{luo2024dynamic,
  title={Dynamic Attribute-guided Few-shot Open-set Network for medical image diagnosis},
  author={Luo, Yiwen and Guo, Xiaoqing and Liu, Li and Yuan, Yixuan},
  journal={Expert Systems with Applications},
  volume={251},
  pages={124098},
  year={2024},
  publisher={Elsevier}
}

@article{yahia2000rough,
  title={Rough neural expert systems},
  author={Yahia, ME and Mahmod, R and Sulaiman, N and Ahmad, F},
  journal={Expert Systems with Applications},
  volume={18},
  number={2},
  pages={87--99},
  year={2000},
  publisher={Elsevier}
}

@article{vellido1999neural,
  title={Neural networks in business: a survey of applications (1992--1998)},
  author={Vellido, Alfredo and Lisboa, Paulo JG and Vaughan, J},
  journal={Expert Systems with applications},
  volume={17},
  number={1},
  pages={51--70},
  year={1999},
  publisher={Elsevier}
}

@inproceedings{sta,
  title={Separate to adapt: Open set domain adaptation via progressive separation},
  author={Liu, Hong and Cao, Zhangjie and Long, Mingsheng and Wang, Jianmin and Yang, Qiang},
  booktitle={Proceedings of the IEEE/CVF Conference on Computer Vision and Pattern Recognition},
  pages={2927--2936},
  year={2019}
}

@inproceedings{ros,
  title={On the effectiveness of image rotation for open set domain adaptation},
  author={Bucci, Silvia and Loghmani, Mohammad Reza and Tommasi, Tatiana},
  booktitle={European Conference on Computer Vision},
  pages={422--438},
  year={2020},
  organization={Springer}
}

@article{torch,
  title={Pytorch: An imperative style, high-performance deep learning library},
  author={Paszke, Adam and Gross, Sam and Massa, Francisco and Lerer, Adam and Bradbury, James and Chanan, Gregory and Killeen, Trevor and Lin, Zeming and Gimelshein, Natalia and Antiga, Luca and others},
  journal={Advances in neural information processing systems},
  volume={32},
  year={2019}
}

@article{dance,
  title={Universal domain adaptation through self supervision},
  author={Saito, Kuniaki and Kim, Donghyun and Sclaroff, Stan and Saenko, Kate},
  journal={Advances in neural information processing systems},
  volume={33},
  pages={16282--16292},
  year={2020}
}

@inproceedings{oh,
  title={Deep hashing network for unsupervised domain adaptation},
  author={Venkateswara, Hemanth and Eusebio, Jose and Chakraborty, Shayok and Panchanathan, Sethuraman},
  booktitle={Proceedings of the IEEE conference on computer vision and pattern recognition},
  pages={5018--5027},
  year={2017}
}

@article{proto_uda,
  title={A prototype-oriented framework for unsupervised domain adaptation},
  author={Tanwisuth, Korawat and Fan, Xinjie and Zheng, Huangjie and Zhang, Shujian and Zhang, Hao and Chen, Bo and Zhou, Mingyuan},
  journal={Advances in Neural Information Processing Systems},
  volume={34},
  pages={17194--17208},
  year={2021}
}

@inproceedings{coal,
  title={Class-imbalanced domain adaptation: An empirical odyssey},
  author={Tan, Shuhan and Peng, Xingchao and Saenko, Kate},
  booktitle={European Conference on Computer Vision},
  pages={585--602},
  year={2020},
  organization={Springer}
}

@inproceedings{unida,
  title={Universal domain adaptation},
  author={You, Kaichao and Long, Mingsheng and Cao, Zhangjie and Wang, Jianmin and Jordan, Michael I},
  booktitle={Proceedings of the IEEE/CVF conference on computer vision and pattern recognition},
  pages={2720--2729},
  year={2019}
}

@inproceedings{isfda,
  title={Imbalanced Source-free Domain Adaptation},
  author={Li, Xinhao and Li, Jingjing and Zhu, Lei and Wang, Guoqing and Huang, Zi},
  booktitle={Proceedings of the 29th ACM International Conference on Multimedia},
  pages={3330--3339},
  year={2021}
}

@article{tao,
  title={Distilling the knowledge in a neural network},
  author={Hinton, Geoffrey and Vinyals, Oriol and Dean, Jeff and others},
  journal={arXiv preprint arXiv:1503.02531},
  volume={2},
  number={7},
  year={2015}
}

@inproceedings{cmu,
  title={Learning to detect open classes for universal domain adaptation},
  author={Fu, Bo and Cao, Zhangjie and Long, Mingsheng and Wang, Jianmin},
  booktitle={European Conference on Computer Vision},
  pages={567--583},
  year={2020},
  organization={Springer}
}

@inproceedings{ova,
  title={Ovanet: One-vs-all network for universal domain adaptation},
  author={Saito, Kuniaki and Saenko, Kate},
  booktitle={Proceedings of the IEEE/CVF International Conference on Computer Vision},
  pages={9000--9009},
  year={2021}
}

@inproceedings{domainnet,
  title={Moment matching for multi-source domain adaptation},
  author={Peng, Xingchao and Bai, Qinxun and Xia, Xide and Huang, Zijun and Saenko, Kate and Wang, Bo},
  booktitle={Proceedings of the IEEE/CVF international conference on computer vision},
  pages={1406--1415},
  year={2019}
}

@inproceedings{visda,
  title={Visda: A synthetic-to-real benchmark for visual domain adaptation},
  author={Peng, Xingchao and Usman, Ben and Kaushik, Neela and Wang, Dequan and Hoffman, Judy and Saenko, Kate},
  booktitle={Proceedings of the IEEE Conference on Computer Vision and Pattern Recognition Workshops},
  pages={2021--2026},
  year={2018}
}

@inproceedings{osbp,
  title={Open set domain adaptation by backpropagation},
  author={Saito, Kuniaki and Yamamoto, Shohei and Ushiku, Yoshitaka and Harada, Tatsuya},
  booktitle={Proceedings of the European Conference on Computer Vision (ECCV)},
  pages={153--168},
  year={2018}
}

@inproceedings{jiang2020implicit,
  title={Implicit class-conditioned domain alignment for unsupervised domain adaptation},
  author={Jiang, Xiang and Lao, Qicheng and Matwin, Stan and Havaei, Mohammad},
  booktitle={International Conference on Machine Learning},
  pages={4816--4827},
  year={2020},
  organization={PMLR}
}

@article{tachet2020domain,
  title={Domain adaptation with conditional distribution matching and generalized label shift},
  author={Tachet des Combes, Remi and Zhao, Han and Wang, Yu-Xiang and Gordon, Geoffrey J},
  journal={Advances in Neural Information Processing Systems},
  volume={33},
  pages={19276--19289},
  year={2020}
}

@inproceedings{tan2020class,
  title={Class-imbalanced domain adaptation: an empirical odyssey},
  author={Tan, Shuhan and Peng, Xingchao and Saenko, Kate},
  booktitle={Computer Vision--ECCV 2020 Workshops: Glasgow, UK, August 23--28, 2020, Proceedings, Part I 16},
  pages={585--602},
  year={2020},
  organization={Springer}
}

@inproceedings{prabhu2021sentry,
  title={Sentry: Selective entropy optimization via committee consistency for unsupervised domain adaptation},
  author={Prabhu, Viraj and Khare, Shivam and Kartik, Deeksha and Hoffman, Judy},
  booktitle={Proceedings of the IEEE/CVF International Conference on Computer Vision},
  pages={8558--8567},
  year={2021}
}

@inproceedings{zendel2022unifying,
  title={Unifying panoptic segmentation for autonomous driving},
  author={Zendel, Oliver and Sch{\"o}rghuber, Matthias and Rainer, Bernhard and Murschitz, Markus and Beleznai, Csaba},
  booktitle={CVPR},
  pages={21351--21360},
  year={2022}
}

@inproceedings{vyas2018out,
  title={Out-of-distribution detection using an ensemble of self supervised leave-out classifiers},
  author={Vyas, Apoorv and Jammalamadaka, Nataraj and Zhu, Xia and Das, Dipankar and Kaul, Bharat and Willke, Theodore L},
  booktitle={ECCV},
  pages={550--564},
  year={2018}
}

@article{ren2019likelihood,
  title={Likelihood ratios for out-of-distribution detection},
  author={Ren, Jie and Liu, Peter J and Fertig, Emily and Snoek, Jasper and Poplin, Ryan and Depristo, Mark and Dillon, Joshua and Lakshminarayanan, Balaji},
  journal={NeurIPS},
  volume={32},
  year={2019}
}

@article{zhou2021step,
  title={Step: Out-of-distribution detection in the presence of limited in-distribution labeled data},
  author={Zhou, Zhi and Guo, Lan-Zhe and Cheng, Zhanzhan and Li, Yu-Feng and Pu, Shiliang},
  journal={NeurIPS},
  volume={34},
  pages={29168--29180},
  year={2021}
}

@article{liu2020energy,
  title={Energy-based out-of-distribution detection},
  author={Liu, Weitang and Wang, Xiaoyun and Owens, John and Li, Yixuan},
  journal={NeurIPS},
  volume={33},
  pages={21464--21475},
  year={2020}
}

@article{sun2021react,
  title={React: Out-of-distribution detection with rectified activations},
  author={Sun, Yiyou and Guo, Chuan and Li, Yixuan},
  journal={NeurIPS},
  volume={34},
  pages={144--157},
  year={2021}
}

@article{lu2023uncertainty,
  title={Uncertainty-Aware Optimal Transport for Semantically Coherent Out-of-Distribution Detection},
  author={Lu, Fan and Zhu, Kai and Zhai, Wei and Zheng, Kecheng and Cao, Yang},
  journal={arXiv preprint arXiv:2303.10449},
  year={2023}
}

@inproceedings{damodaran2018deepjdot,
  title={Deepjdot: Deep joint distribution optimal transport for unsupervised domain adaptation},
  author={Damodaran, Bharath Bhushan and Kellenberger, Benjamin and Flamary, R{\'e}mi and Tuia, Devis and Courty, Nicolas},
  booktitle={Proceedings of the European conference on computer vision (ECCV)},
  pages={447--463},
  year={2018}
}

@inproceedings{ganin2015unsupervised,
  title={Unsupervised domain adaptation by backpropagation},
  author={Ganin, Yaroslav and Lempitsky, Victor},
  booktitle={International conference on machine learning},
  pages={1180--1189},
  year={2015},
  organization={PMLR}
}

@inproceedings{long2015learning,
  title={Learning transferable features with deep adaptation networks},
  author={Long, Mingsheng and Cao, Yue and Wang, Jianmin and Jordan, Michael},
  booktitle={International conference on machine learning},
  pages={97--105},
  year={2015},
  organization={PMLR}
}

@inproceedings{long2017deep,
  title={Deep transfer learning with joint adaptation networks},
  author={Long, Mingsheng and Zhu, Han and Wang, Jianmin and Jordan, Michael I},
  booktitle={International conference on machine learning},
  pages={2208--2217},
  year={2017},
  organization={PMLR}
}

@article{tzeng2014deep,
  title={Deep domain confusion: Maximizing for domain invariance},
  author={Tzeng, Eric and Hoffman, Judy and Zhang, Ning and Saenko, Kate and Darrell, Trevor},
  journal={arXiv preprint arXiv:1412.3474},
  year={2014}
}

@inproceedings{chen2019temporal,
  title={Temporal attentive alignment for large-scale video domain adaptation},
  author={Chen, Min-Hung and Kira, Zsolt and AlRegib, Ghassan and Yoo, Jaekwon and Chen, Ruxin and Zheng, Jian},
  booktitle={Proceedings of the IEEE/CVF International Conference on Computer Vision},
  pages={6321--6330},
  year={2019}
}

@inproceedings{chen2020action,
  title={Action segmentation with joint self-supervised temporal domain adaptation},
  author={Chen, Min-Hung and Li, Baopu and Bao, Yingze and AlRegib, Ghassan and Kira, Zsolt},
  booktitle={Proceedings of the IEEE/CVF Conference on Computer Vision and Pattern Recognition},
  pages={9454--9463},
  year={2020}
}

@inproceedings{munro2020multi,
  title={Multi-modal domain adaptation for fine-grained action recognition},
  author={Munro, Jonathan and Damen, Dima},
  booktitle={Proceedings of the IEEE/CVF conference on computer vision and pattern recognition},
  pages={122--132},
  year={2020}
}

@inproceedings{choi2020shuffle,
  title={Shuffle and attend: Video domain adaptation},
  author={Choi, Jinwoo and Sharma, Gaurav and Schulter, Samuel and Huang, Jia-Bin},
  booktitle={Computer Vision--ECCV 2020: 16th European Conference, Glasgow, UK, August 23--28, 2020, Proceedings, Part XII 16},
  pages={678--695},
  year={2020},
  organization={Springer}
}

@article{van2008visualizing,
  title={Visualizing data using t-SNE.},
  author={Van der Maaten, Laurens and Hinton, Geoffrey},
  journal={Journal of machine learning research},
  volume={9},
  number={11},
  year={2008}
}

@article{vernekar2019out,
  title={Out-of-distribution detection in classifiers via generation},
  author={Vernekar, Sachin and Gaurav, Ashish and Abdelzad, Vahdat and Denouden, Taylor and Salay, Rick and Czarnecki, Krzysztof},
  journal={arXiv preprint arXiv:1910.04241},
  year={2019}
}

@article{ru2023imbalanced,
  title={Imbalanced open set domain adaptation via moving-threshold estimation and gradual alignment},
  author={Ru, Jinghan and Tian, Jun and Xiao, Chengwei and Li, Jingjing and Shen, Heng Tao},
  journal={IEEE Transactions on Multimedia},
  year={2023},
  publisher={IEEE}
}

@inproceedings{yang2022multi,
  title={On multi-domain long-tailed recognition, imbalanced domain generalization and beyond},
  author={Yang, Yuzhe and Wang, Hao and Katabi, Dina},
  booktitle={European Conference on Computer Vision},
  pages={57--75},
  year={2022},
  organization={Springer}
}

@inproceedings{sun2022exploiting,
  title={Exploiting mixed unlabeled data for detecting samples of seen and unseen out-of-distribution classes},
  author={Sun, Yi-Xuan and Wang, Wei},
  booktitle={Proceedings of the AAAI Conference on Artificial Intelligence},
  volume={36},
  number={8},
  pages={8386--8394},
  year={2022}
}

@inproceedings{li2020background,
  title={Background data resampling for outlier-aware classification},
  author={Li, Yi and Vasconcelos, Nuno},
  booktitle={Proceedings of the IEEE/CVF Conference on Computer Vision and Pattern Recognition},
  pages={13218--13227},
  year={2020}
}

@inproceedings{kang2019contrastive,
  title={Contrastive adaptation network for unsupervised domain adaptation},
  author={Kang, Guoliang and Jiang, Lu and Yang, Yi and Hauptmann, Alexander G},
  booktitle={Proceedings of the IEEE/CVF conference on computer vision and pattern recognition},
  pages={4893--4902},
  year={2019}
}

@inproceedings{melas2021pixmatch,
  title={Pixmatch: Unsupervised domain adaptation via pixelwise consistency training},
  author={Melas-Kyriazi, Luke and Manrai, Arjun K},
  booktitle={Proceedings of the IEEE/CVF Conference on Computer Vision and Pattern Recognition},
  pages={12435--12445},
  year={2021}
}

@inproceedings{peng2020domain2vec,
  title={Domain2vec: Domain embedding for unsupervised domain adaptation},
  author={Peng, Xingchao and Li, Yichen and Saenko, Kate},
  booktitle={Computer Vision--ECCV 2020: 16th European Conference, Glasgow, UK, August 23--28, 2020, Proceedings, Part VI 16},
  pages={756--774},
  year={2020},
  organization={Springer}
}

@article{fort2021exploring,
  title={Exploring the limits of out-of-distribution detection},
  author={Fort, Stanislav and Ren, Jie and Lakshminarayanan, Balaji},
  journal={Advances in Neural Information Processing Systems},
  volume={34},
  pages={7068--7081},
  year={2021}
}

@inproceedings{sun2022dice,
  title={Dice: Leveraging sparsification for out-of-distribution detection},
  author={Sun, Yiyou and Li, Yixuan},
  booktitle={European Conference on Computer Vision},
  pages={691--708},
  year={2022},
  organization={Springer}
}

@inproceedings{du2020fine,
  title={Fine-grained visual classification via progressive multi-granularity training of jigsaw patches},
  author={Du, Ruoyi and Chang, Dongliang and Bhunia, Ayan Kumar and Xie, Jiyang and Ma, Zhanyu and Song, Yi-Zhe and Guo, Jun},
  booktitle={European Conference on Computer Vision},
  pages={153--168},
  year={2020},
  organization={Springer}
}

@inproceedings{zhang2022tip,
  title={Tip-adapter: Training-free adaption of clip for few-shot classification},
  author={Zhang, Renrui and Zhang, Wei and Fang, Rongyao and Gao, Peng and Li, Kunchang and Dai, Jifeng and Qiao, Yu and Li, Hongsheng},
  booktitle={European Conference on Computer Vision},
  pages={493--510},
  year={2022},
  organization={Springer}
}

@inproceedings{rastegari2016xnor,
  title={Xnor-net: Imagenet classification using binary convolutional neural networks},
  author={Rastegari, Mohammad and Ordonez, Vicente and Redmon, Joseph and Farhadi, Ali},
  booktitle={European conference on computer vision},
  pages={525--542},
  year={2016},
  organization={Springer}
}

@inproceedings{cho2022towards,
  title={Towards accurate open-set recognition via background-class regularization},
  author={Cho, Wonwoo and Choo, Jaegul},
  booktitle={European Conference on Computer Vision},
  pages={658--674},
  year={2022},
  organization={Springer}
}

@inproceedings{neal2018open,
  title={Open set learning with counterfactual images},
  author={Neal, Lawrence and Olson, Matthew and Fern, Xiaoli and Wong, Weng-Keen and Li, Fuxin},
  booktitle={Proceedings of the European Conference on Computer Vision (ECCV)},
  pages={613--628},
  year={2018}
}

@inproceedings{de2021impact,
  title={Impact of colour on robustness of deep neural networks},
  author={De, Kanjar and Pedersen, Marius},
  booktitle={Proceedings of the IEEE/CVF international conference on computer vision},
  pages={21--30},
  year={2021}
}

@inproceedings{reimers2020determining,
  title={Determining the relevance of features for deep neural networks},
  author={Reimers, Christian and Runge, Jakob and Denzler, Joachim},
  booktitle={European Conference on Computer Vision},
  pages={330--346},
  year={2020},
  organization={Springer}
}

@inproceedings{venkateswara2017deep,
  title={Deep hashing network for unsupervised domain adaptation},
  author={Venkateswara, Hemanth and Eusebio, Jose and Chakraborty, Shayok and Panchanathan, Sethuraman},
  booktitle={Proceedings of the IEEE conference on computer vision and pattern recognition},
  pages={5018--5027},
  year={2017}
}

@article{long2016unsupervised,
  title={Unsupervised domain adaptation with residual transfer networks},
  author={Long, Mingsheng and Zhu, Han and Wang, Jianmin and Jordan, Michael I},
  journal={Advances in neural information processing systems},
  volume={29},
  year={2016}
}

@inproceedings{serrainput,
  title={Input Complexity and Out-of-distribution Detection with Likelihood-based Generative Models},
  author={Serr{\`a}, Joan and {\'A}lvarez, David and G{\'o}mez, Vicen{\c{c}} and Slizovskaia, Olga and N{\'u}{\~n}ez, Jos{\'e} F and Luque, Jordi},
  booktitle={ICLR},
  year={2019}
}

@article{lee2018simple,
  title={A simple unified framework for detecting out-of-distribution samples and adversarial attacks},
  author={Lee, Kimin and Lee, Kibok and Lee, Honglak and Shin, Jinwoo},
  journal={NeurIPS},
  volume={31},
  year={2018}
}

@inproceedings{yang2022out,
  title={Out-of-distribution detection with semantic mismatch under masking},
  author={Yang, Yijun and Gao, Ruiyuan and Xu, Qiang},
  booktitle={ECCV},
  pages={373--390},
  year={2022},
  organization={Springer}
}

@inproceedings{yu2019unsupervised,
  title={Unsupervised out-of-distribution detection by maximum classifier discrepancy},
  author={Yu, Qing and Aizawa, Kiyoharu},
  booktitle={ICCV},
  pages={9518--9526},
  year={2019}
}

@inproceedings{mohseni2020self,
  title={Self-supervised learning for generalizable out-of-distribution detection},
  author={Mohseni, Sina and Pitale, Mandar and Yadawa, JBS and Wang, Zhangyang},
  booktitle={AAAI},
  volume={34},
  number={04},
  pages={5216--5223},
  year={2020}
}

@inproceedings{zaeemzadeh2021out,
  title={Out-of-distribution detection using union of 1-dimensional subspaces},
  author={Zaeemzadeh, Alireza and Bisagno, Niccolo and Sambugaro, Zeno and Conci, Nicola and Rahnavard, Nazanin and Shah, Mubarak},
  booktitle={CVPR},
  pages={9452--9461},
  year={2021}
}

@inproceedings{hsu2020generalized,
  title={Generalized odin: Detecting out-of-distribution image without learning from out-of-distribution data},
  author={Hsu, Yen-Chang and Shen, Yilin and Jin, Hongxia and Kira, Zsolt},
  booktitle={CVPR},
  pages={10951--10960},
  year={2020}
}

@inproceedings{ming2023exploit,
  title={How to Exploit Hyperspherical Embeddings for Out-of-Distribution Detection?},
  author={Ming, Yifei and Sun, Yiyou and Dia, Ousmane and Li, Yixuan},
  booktitle={The Eleventh ICLR},
  year={2023}
}

@inproceedings{liang2018enhancing,
  title={Enhancing the reliability of out-of-distribution image detection in neural networks},
  author={Liang, Shiyu and Li, Yixuan and Srikant, R},
  booktitle={ICLR},
  year={2018}
}

@inproceedings{turrisi2022multi,
  title={Multi-source domain adaptation via weighted joint distributions optimal transport},
  author={Turrisi, Rosanna and Flamary, R{\'e}mi and Rakotomamonjy, Alain and Pontil, Massimiliano},
  booktitle={Uncertainty in Artificial Intelligence},
  pages={1970--1980},
  year={2022},
  organization={PMLR}
}

@inproceedings{nguyen2021most,
  title={Most: Multi-source domain adaptation via optimal transport for student-teacher learning},
  author={Nguyen, Tuan and Le, Trung and Zhao, He and Tran, Quan Hung and Nguyen, Truyen and Phung, Dinh},
  booktitle={Uncertainty in Artificial Intelligence},
  pages={225--235},
  year={2021},
  organization={PMLR}
}

@inproceedings{hendrycksbaseline,
  title={A Baseline for Detecting Misclassified and Out-of-Distribution Examples in Neural Networks},
  author={Hendrycks, Dan and Gimpel, Kevin},
  year={2016},
  booktitle={ICLR}
}

@inproceedings{lee2018hierarchical,
  title={Hierarchical novelty detection for visual object recognition},
  author={Lee, Kibok and Lee, Kimin and Min, Kyle and Zhang, Yuting and Shin, Jinwoo and Lee, Honglak},
  booktitle={CVPR},
  pages={1034--1042},
  year={2018}
}

@inproceedings{lee2018training,
  title={Training confidence-calibrated classifiers for detecting out-of-distribution samples},
  author={Lee, Kimin and Lee, Honglak and Lee, Kibok and Shin, Jinwoo},
  booktitle={ICLR},
  year={2018}
}

@inproceedings{techapanurak2020hyperparameter,
  title={Hyperparameter-free out-of-distribution detection using cosine similarity},
  author={Techapanurak, Engkarat and Suganuma, Masanori and Okatani, Takayuki},
  booktitle={ACCV},
  year={2020}
}

@article{kirichenko2020normalizing,
  title={Why normalizing flows fail to detect out-of-distribution data},
  author={Kirichenko, Polina and Izmailov, Pavel and Wilson, Andrew G},
  journal={NeurIPS},
  volume={33},
  pages={20578--20589},
  year={2020}
}

@inproceedings{zhou2022rethinking,
  title={Rethinking reconstruction autoencoder-based out-of-distribution detection},
  author={Zhou, Yibo},
  booktitle={CVPR},
  pages={7379--7387},
  year={2022}
}

@inproceedings{hendrycks2017baseline,
  title={A Baseline for Detecting Misclassified and Out-of-Distribution Examples in Neural Networks},
  author={Hendrycks, Dan and Gimpel, Kevin},
  booktitle={International Conference on Learning Representations},
  year={2017}
}

@inproceedings{nguyen2022cycle,
  title={Cycle class consistency with distributional optimal transport and knowledge distillation for unsupervised domain adaptation},
  author={Nguyen, Tuan and Nguyen, Van and Le, Trung and Zhao, He and Tran, Quan Hung and Phung, Dinh},
  booktitle={Uncertainty in Artificial Intelligence},
  pages={1519--1529},
  year={2022},
  organization={PMLR}
}

@inproceedings{sicilia2022pac,
  title={PAC-bayesian domain adaptation bounds for multiclass learners},
  author={Sicilia, Anthony and Atwell, Katherine and Alikhani, Malihe and Hwang, Seong Jae},
  booktitle={Uncertainty in Artificial Intelligence},
  pages={1824--1834},
  year={2022},
  organization={PMLR}
}

@inproceedings{saito2021ovanet,
  title={Ovanet: One-vs-all network for universal domain adaptation},
  author={Saito, Kuniaki and Saenko, Kate},
  booktitle={Proceedings of the ieee/cvf international conference on computer vision},
  pages={9000--9009},
  year={2021}
}

\end{document}